\documentclass{article}

\PassOptionsToPackage{numbers,sort&compress}{natbib}
\usepackage[dblblindworkshop,final]{neurips_2026}

\usepackage[utf8]{inputenc}
\usepackage[T1]{fontenc}
\usepackage{lmodern}
\usepackage[colorlinks=true,linkcolor=blue,citecolor=blue,urlcolor=blue]{hyperref}
\usepackage{url}
\usepackage{booktabs}
\usepackage{tabularx}
\usepackage{amsfonts, amsmath, amssymb}
\usepackage{nicefrac}
\usepackage{microtype}
\usepackage{graphicx}
\usepackage{multirow}
\usepackage{enumitem}
\usepackage{float}
\usepackage[table]{xcolor}
\usepackage{caption}
\newcommand{\Text}{T_{\mathrm{ext}}}
\newcommand{\vext}{v_{\mathrm{ext}}}

\newcommand{\degC}{${}^{\circ}$C}

\title{Which alloy composition,\\what process parameters?\\
Inferring the recipe from optimized metallic~microstructure and texture}

\author{%
  Mahish K. Guru$^{1,2}$ \quad Jan Bohlen$^{1}$ \quad Louam Lemjid$^{1}$
  \quad Marius Tacke$^{3}$\\
  \textbf{Roland Aydin}$^{4,5}$ \quad \textbf{Noomane Ben Khalifa}$^{1,2}$\\[0.4em]
  \small $^{1}$Institute of Material and Process Design, Helmholtz-Zentrum Hereon, Germany\\
  \small $^{2}$Institute of Production Technology and Systems, Leuphana University
  L\"uneburg, Germany\\
  \small $^{3}$Institute of Material Systems Modeling, Helmholtz-Zentrum Hereon, Germany\\
  \small $^{4}$AI for Physical Systems, German Research Center for Artificial
  Intelligence (DFKI), Germany\\
  \small $^{5}$Saarland University, Germany
}

\begin{document}

\maketitle

\begin{abstract}
The mechanical properties of a metallic alloy are set by its microstructure and texture: the size and shape of its grains and the orientation of their crystals. That structure is in turn set by a recipe, the alloy composition together with the processing parameters. Alloy development runs this chain forwards, tuning the structure until a target property is met. Running it backwards, from an optimized structure to the recipe that would produce it, still relies on expert knowledge. We ask whether this backwards step can be learned. On an in-house dataset of 107 magnesium alloy extrusion conditions across 14 alloys, each with optical micrographs and an X-ray texture measurement, we compare three descriptors of microstructure and texture: conventional grain and texture statistics, a vision embedding from a pretrained image encoder, and a graph neural network on the grain network. Each is paired with prediction heads for two tasks: the alloy composition given the process (Task A), and the process parameters given the composition (Task B). Under 5-fold cross-validation, the conventional descriptors identify the correct alloy for 65\% of held-out conditions, against 17\% for always guessing the most common alloy, while the learned embeddings stay below 30\%. The process parameters are recoverable but noisier: compared with using the composition alone, the microstructure roughly halves the temperature error. Because only a few alloys were cast and only a few press settings were used, both answers are discrete, and heads that pick from these known options, while respecting their order, worked better than heads that predict a free value.
\end{abstract}

\section{Introduction}
\label{sec:intro}

\begin{figure}[t]
  \centering
  \includegraphics[width=0.99\linewidth]{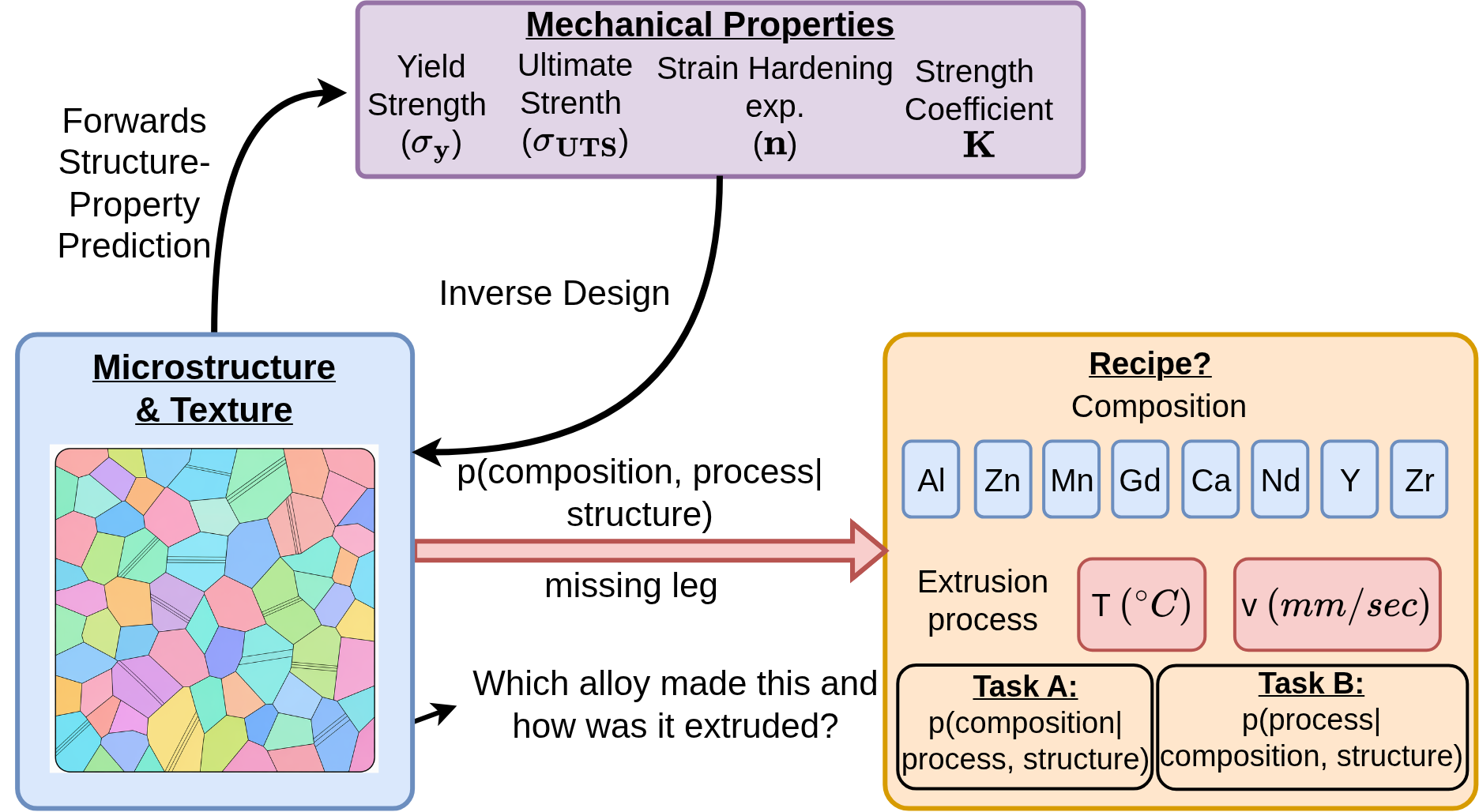}
  \caption{The forward leg predicts
  mechanical properties from microstructure and texture; inverse design (the
  design leg) goes from a target property back to a structure. The missing
  recipe leg (red arrow), studied here, infers the recipe from that structure.
  We evaluate it as two conditional tasks: Task A predicts the composition
  given the structure and the known process, and Task B the process given the
  structure and the known composition.}
  \label{fig:missing_leg}
\end{figure}

Alloy development is a chain of two links. A \emph{recipe}, the alloy
composition together with the processing parameters, produces a microstructure
and texture; that structure in turn sets the mechanical properties. Research
has concentrated on the second link, and travels it in both directions. The
\textbf{forward leg} goes from structure to property: for extruded magnesium
alloys, statistical microstructure and texture descriptors predict
strain-hardening behavior within a few percent
\cite{Guru2025,INDURKAR2020102762,PAULSON2017428}. The \textbf{design leg}
goes from property back to structure: a mature literature describes how to
tailor grains, boundaries and crystallographic orientation for a desired
response \cite{Kambayashi31122025,Lin2026,Wu2021}, and generative latent
optimization can propose microstructures for a target stress-strain curve
\cite{KUSAMPUDI2023103776}. Together these two legs deliver an optimized
structure, and there the chain stops: the first link is rarely run backwards.
That step is the \textbf{recipe leg}, from an optimized structure to the
recipe that would produce it, and it still leans on expert knowledge:
\emph{given only the structure, which alloy composition and what process
parameters made it?} This paper studies the recipe leg
(Figure~\ref{fig:missing_leg}).

In automated characterization pipelines and self-driving laboratories, the
recipe leg serves quality control, anomaly detection and the planning of the
next experiment \cite{Tom2024,Lee2026,Adesiji,Kalinin2021,Kalinin2022DeepLF}.
Such a laboratory proposes a recipe, runs it and images the result. Nothing in
that loop checks that the structure under the microscope is the one the recipe
should have produced. A model of the recipe leg supplies that check from
images already being collected, and can flag a mislabelled sample, a drifting
press or a repeated condition. It is also the first question in reverse
engineering a legacy or competitor material. The difficulty in practice is
the messiness of real experiments: campaigns are small, heterogeneous, pooled
across sources and thinly sampled in every variable. We work in that regime
instead of avoiding it with simulated data.

Our data come from the measured campaign behind our earlier forward-leg study
\cite{Guru2025}, which used microstructure and texture statistics of extruded
magnesium alloys to predict their strain-hardening parameters. The campaign
covers 107 extrusion conditions of wrought magnesium alloys, a hexagonal
close-packed (HCP) metal,
over 14 alloy classes, with optical micrographs and an X-ray texture
measurement for each condition. Both halves of the recipe look continuous but
are not. The composition is a weight percentage for each of eight alloying
elements (Al, Zn, Mn, Ce, Gd, Ca, Nd, Y), but only 14 alloys were cast, so the
answer is always one of 14 known compositions. The process behaves the same
way: extrusion temperature and ram speed could take any value, but the press
was run at a short list of settings. Both answers are therefore discrete, and
this shapes the rest of the paper.

We compare three descriptors of microstructure and texture. The first is
conventional grain and texture statistics, a domain-designed pipeline computed
directly from the micrographs and the measured texture. The second is an
embedding from the image encoder of our Co-PiLOT work
\cite{guru2026copilotconstrainedphysicsinformedlatent}, pretrained on about
80k orientation maps and used frozen. The third is a graph neural network on
the network of grains and their boundaries. Each descriptor is passed to
prediction heads, the models that turn it into an answer, which either respect
or ignore the discrete answer. There are two tasks: Task A, the alloy
composition, and Task B, the process parameters. The
question we put to each descriptor is whether it is \emph{compatible} with the
recipe leg: does this way of describing an optimized microstructure and
texture support recovering the recipe behind it? What this paper offers:

\begin{enumerate}[leftmargin=1.4em,itemsep=1pt]
  \item We state the recipe leg as two tasks that a model can be scored on,
  composition given the known process and process given the known
  composition, and pose them on a measured campaign rather than on simulated
  data.
  \item We compare conventional grain and texture statistics, a frozen vision
  embedding and a grain-graph network on the same folds, heads and metrics,
  and state exactly what each of them receives.
  \item We give prediction heads that decode onto the short list of alloys and
  press settings that exist, and compare each against the same evidence used
  without that constraint and against baselines that see no microstructure.
  \item We report, next to the headline error, metrics that reveal when a head
  scores well for the wrong reason, and show heads from our own search that
  these metrics exposed.
\end{enumerate}

\section{Related work}
\label{sec:related}

Most data-driven work on microstructure runs the forward direction, from
structure to property, using spatial correlations, reduced statistics or
learned features \cite{Kalidindi-MKS-2015,Guru2025,INDURKAR2020102762,PAULSON2017428}.
Inverse design inverts that map to propose a structure for a target property
\cite{KUSAMPUDI2023103776,guru2026copilotconstrainedphysicsinformedlatent}, and
reconstruction methods generate structures that match measured statistics
\cite{bostanabad2018computational}. Process-structure linkages are also
mostly studied forwards, predicting the structure a process will produce,
often on simulated data \cite{popova2017process}. Closest to our question
are computer-vision studies that classify micrographs by their origin.
Generic image features can sort microstructure images into classes
\cite{decost2015computer}, and image-texture representations of
ultrahigh-carbon steel separate micrographs by their annealing conditions
\cite{decost2017exploring}. These studies infer one processing label for
one material system. We ask for the full recipe, composition and process,
across 14 alloys, and score the answer against the discrete set of recipes
that were actually run.

Learned representations of microstructure range from CNN features pretrained
on natural images \cite{gatys2015texture}\cite{holm2020overview} and encoders
pretrained on large microscopy collections \cite{stuckner2022microstructure}
to graph neural networks over grains and their boundaries
\cite{dai2021graph} and generative latent spaces for inverse design
\cite{guru2026copilotconstrainedphysicsinformedlatent}. Most of them are
evaluated where data are plentiful or simulated. Experimental campaigns
instead have tens to hundreds of conditions, with labels drawn from a short
list of settings \cite{Kalinin2021,Tom2024}. This paper sits between the two
lines of work. It compares a domain-designed descriptor with two learned ones
on such a campaign, and asks which descriptor, combined with which kind of
prediction head, recovers the recipe.

\section{Data: sparse by construction}
\label{sec:data}

The dataset aggregates measured extrusion campaigns
of wrought magnesium alloys: AZ31 from a doctoral thesis
\cite{nienaber_einfluss_der_2023} and a processing-parameter study
\cite{Nienaber2019}, the Zn-bearing alloys Z1, ZX10 and ZNd10 from a Zn-bearing
extrusion study \cite{CANOCASTILLO2020139527,Nienaber2019}, ME21 from an
extrusion-parameter study \cite{KurzG}, and the Mg-Gd family Mg-2/5/10Gd with
0, 0.5, or 1 wt\% Mn micro-additions from Mg-Gd and Mg-Gd-Mn extrusion studies
\cite{Harmuth,cryst12081036}. Each of the 107 conditions combines a base alloy (composition
in wt\% over the eight alloying elements Al, Zn, Mn, Ce, Gd, Ca, Nd, Y), an extrusion
temperature $\Text \in [200, 500]$\,°C, and a ram speed $\vext \in [0.5,
7.5]$\,mm/s. Per condition we have optical micrographs and X-ray pole figures reduced to an orientation
distribution function (ODF).

\textbf{Two kinds of sparsity.} The labels of the composition task are
sparse: 14 alloys are spread thinly over eight possible alloying elements,
most elements are absent from most alloys, and Y appears in a single alloy.
The labels of the process parameter task are sparse and also confounded: each
alloy family was extruded on its own temperature and speed grid, so
process and composition are correlated by experimental design. Any model can
exploit that correlation, so we also ask whether the microstructure adds
information beyond it. Every result is therefore compared with a baseline
that sees the known half of the recipe but no microstructure.

\textbf{Two tasks.} We split the recipe leg into two tasks.
\textbf{Task A, the composition task:} given the
structure representation plus the known process $(\Text, \log \vext)$ and an
extrusion-ratio-type flag, predict the element wt\% vector.
\textbf{Task B, the process parameter task:} given the representation plus the
known composition
and the same flag, predict $(\Text, \vext)$. Each task receives the half of
the recipe that a user would usually already know.

Each condition has many images, so a random split by image would put images
of the same condition in both the training and the test set; our fusion head
(Section~\ref{sec:heads}) then scores 100\% alloy top-1. All results
therefore use condition-grouped 5-fold cross-validation: every condition, with
all its images, sits in exactly one held-out fold, and each test fold is used
once. A validation part of each training fold selects hyperparameters, and
every other choice is made on out-of-fold (OOF) predictions, that is, on
training conditions held out inside the training fold.

\section{Method: microstructure and texture descriptors}
\label{sec:repr}

\begin{figure}[t]
  \centering
  \includegraphics[width=\linewidth]{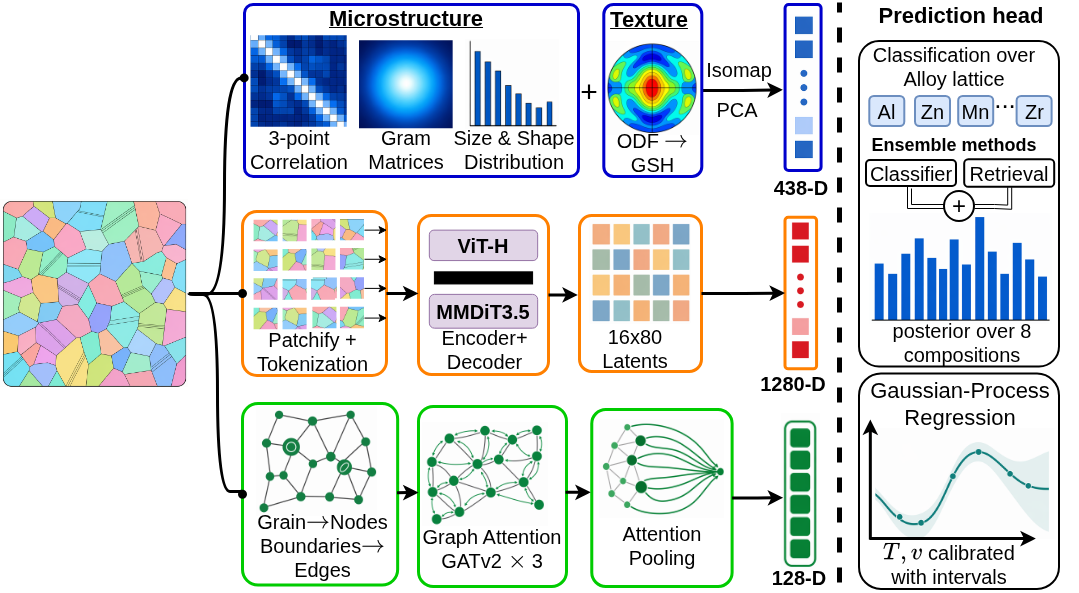}
  \caption{Three descriptor branches with shared prediction heads.
  \textbf{Top (blue):} conventional statistics of the micrograph and the
  measured ODF, 438-D. \textbf{Middle (orange):} the frozen Co-PiLOT ViT-H
  encoder reading an orientation-coloured RVE map, 1280-D
  (Figure~\ref{fig:vision}). \textbf{Bottom (green):} grain graph of the same
  RVE through a graph attention encoder, 128-D. Right: the heads, shared by
  all branches. Exact inputs in Table~\ref{tab:inputs}.}
  \label{fig:pipeline}
\end{figure}

Figure~\ref{fig:pipeline} shows the three descriptor branches and
Table~\ref{tab:inputs} lists exactly what each receives. They cover the
range of options a practitioner realistically has, from conventional
statistics to vision embeddings, and the benchmark asks which of them is
compatible with the recipe leg at 107 conditions.

\textbf{What each branch sees.} All three branches start from the same
measurements, one optical micrograph at a time plus the measured ODF of its
condition, but read them in different forms. The conventional branch computes
its statistics directly from the micrograph, its grain mask and the ODF. The
two learned branches read a representative volume element (RVE) built for
each micrograph in DREAM.3D \cite{groeber2014}: grain sizes and aspect ratios
follow that micrograph's measured distributions, and grain orientations are
sampled from the condition's measured ODF. The RVE is rendered as an
orientation-coloured map, in which each pixel's colour encodes the local
crystal orientation as in an EBSD map, for the vision encoder, and segmented
into a grain graph for the GNN. All three branches therefore receive the
texture. They differ in how they see the grain arrangement: the conventional
branch measures it on the micrograph itself, while the learned branches see it
as reproduced by the RVE from the measured statistics.

\begin{table}[t]
  \caption{Inputs per branch. Every branch also receives the known half of
  the recipe: $\Text$, $\log\vext$ and a two-column extrusion-ratio flag in
  Task A, or the eight element wt\% and the flag in Task B. Last column: input
  width at the kNN, fusion and GBDT heads / the reranker's second classifier /
  the ARD GP / the joint-grid GP (Section~\ref{sec:heads}). GSH =
  generalized spherical harmonics.}
  \label{tab:inputs}
  \centering\footnotesize
  \begin{tabularx}{\linewidth}{@{}lXXl@{}}
    \toprule
    branch & image input & texture enters as & width at the heads \\
    \midrule
    conv. & optical micrograph and its grain mask & GSH coefficients of the ODF & 438 / 5 blocks / 32 / 8 \\
    vision & orientation-coloured RVE map & orientations sampled from the ODF & 1280 / $\le$128 / 32 / 8 \\
    gnn & grain graph of the same RVE & per-grain Euler angles from the ODF & 128 / 128 / 32 / 8 \\
    \bottomrule
  \end{tabularx}
\end{table}

\textbf{Conventional grain and texture statistics (438-D).} This pipeline is
taken from our forward-leg study in Acta Materialia \cite{Guru2025}, where the
same statistics predicted strain-hardening parameters within a few percent.
Each image contributes:
PCA-reduced 3-point spatial correlations (120-D), Isomap-reduced Gram matrices
of pretrained CNN activations (200-D), aspect-ratio and equivalent-diameter
distributions (58-D), and from the ODF, Isomap-reduced generalized spherical
harmonics coefficients (60-D). These descriptors are cheap, deterministic, and
built from domain knowledge of what distinguishes one extruded
microstructure from another \cite{Kalidindi-MKS-2015,markley2007,tenenbaum2000isomap,gatys2015texture,gatys2016style}.

\textbf{Vision embedding (1280-D).} The middle branch reuses the encoder of
our Co-PiLOT work \cite{guru2026copilotconstrainedphysicsinformedlatent},
which learns a latent space of orientation maps and optimizes in it to design
microstructures for a target property. There, a ViT-H/14 trunk
\cite{dosovitskiy2021image} initialized from natural-image CLIP weights
\cite{radford2021clip} was trained on about 80k EBSD-derived microstructures,
with its upper 16 blocks fine-tuned, and an attention pooler reads it out into
16 latent tokens. A Stable Diffusion 3.5 MMDiT decoder \cite{esser2024sd3}
reconstructs the image from those tokens, so the latent is shaped by
reconstruction rather than by any recipe label
\cite{peebles2023dit,lipman2023flowmatching}. In this paper the whole encoder
is frozen: nothing is fine-tuned on the 107 conditions, and the
$16 \times 80$ tokens are flattened to the 1280-D embedding
(Figure~\ref{fig:vision} in Appendix~\ref{app:descriptors}). The question is
whether a latent learned for reconstruction transfers to the recipe leg.

\begin{figure}[t]
  \centering
  \includegraphics[width=\linewidth]{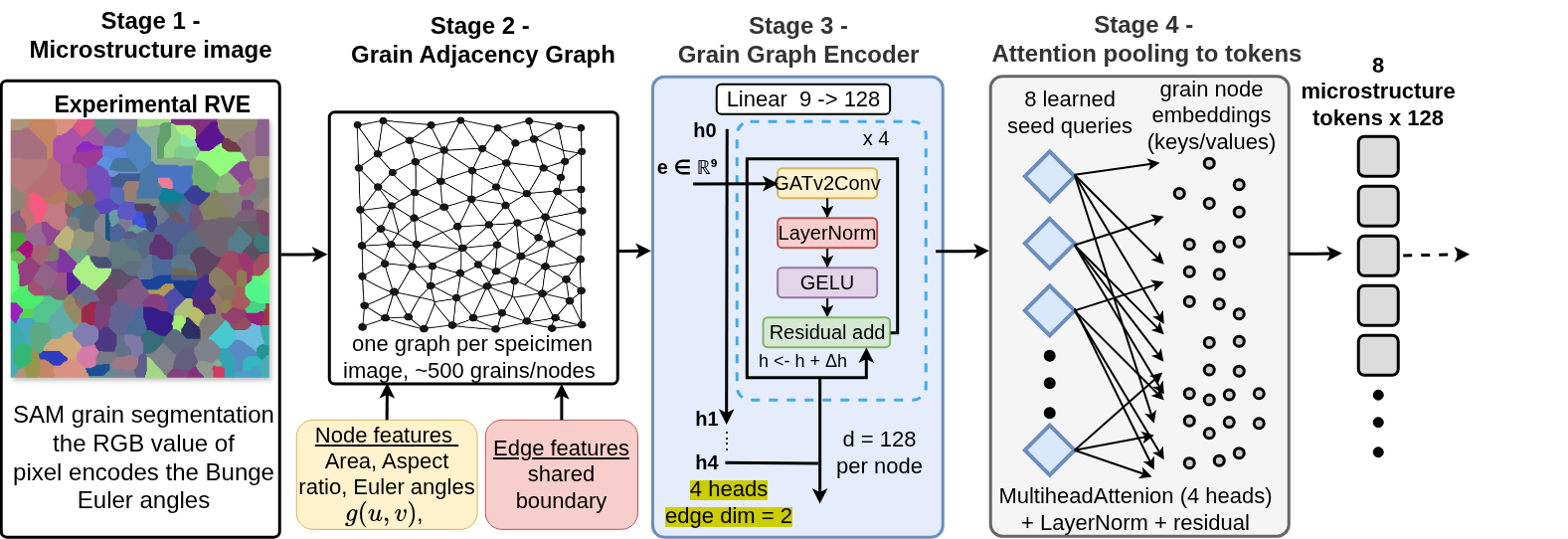}
  \caption{Grain-graph encoder. Stage 1: segmented grain map of the RVE.
  Stage 2: grain adjacency graph, roughly 500 grains per image. Stage 3: four
  GATv2 blocks with residual connections. Stage 4: attention pooling to 8
  microstructure tokens.}
  \label{fig:gnn}
\end{figure}

\textbf{Grain-graph embedding (128-D).}
The third branch turns each RVE into a graph whose nodes are
grains and whose edges are shared boundaries. Node features
cover grain geometry and orientation, and edge features cover shared-boundary
length and misorientation angle \cite{randle2001grain}. A graph attention
encoder \cite{brody2022how, velickovic2018graph} followed by attention pooling
\cite{vaswani2017attention} produces 8 microstructure tokens per image, which
are mean-pooled to the 128-D embedding used in the benchmark. Unlike the
other two pipelines, this encoder was designed and trained for this study. It
is trained once with a flow-matching objective \cite{lipman2023flowmatching}
and then frozen, so each fold reuses cached embeddings. Figure~\ref{fig:gnn}
shows the four stages and Appendix~\ref{app:descriptors} the layer sizes.

\section{Method: prediction heads}
\label{sec:heads}

A descriptor turns a micrograph into a vector; a \emph{prediction head} turns
that vector into an answer to one of the two tasks. Section~\ref{sec:data}
showed that both answers are discrete. We therefore build heads that choose
from the known alloys or settings, which we call the catalogue, and compare
each with a head that sees the same input but may answer anywhere in a
continuous range. All heads see
the same descriptor input on the same folds. Two baselines that see no
microstructure anchor each task: the most frequent training answer, and a
classifier that sees only the known half of the recipe. Appendix~\ref{app:heads}
gives the full specifications, and Appendix~\ref{app:metrics} the metric
equations and the reason for each baseline.

\textbf{Task A: choosing among known alloys.} We predict which of the 14
alloys a condition belongs to and report the nominal composition of that alloy
as the predicted element weight percentages. Image-level probabilities are
averaged within a condition first, so every condition contributes one
prediction regardless of how many micrographs it has. Three heads of
increasing structure are compared. \textbf{Retrieval} lets the alloy labels of
the nearest training images in descriptor space vote, the natural baseline
when the answer is a catalogue entry \cite{cover1967nearest,
wang2014learning}. \textbf{Condition-balanced fusion} averages that vote with
a 14-way gradient-boosted classifier trained with equal weight per condition,
so conditions with many images do not dominate
\cite{prokhorenkova2018catboost,friedman2001greedy, chen2016xgboost}. The
\textbf{reranker}, our main head, blends the fusion with a second classifier
chosen to make different mistakes: one logistic classifier per descriptor
block on the conventional branch, and an FT-Transformer on the two learned
branches, which have no block structure \cite{hastie2009elements,
gorishniy2021revisiting, badaro2023transformers}. Its blend weight is selected
on OOF predictions, and only weights that do not worsen the guard metrics
below are eligible \cite{wolpert1992stacked, varma2006bias}. When none
qualifies, the fusion is kept unchanged, so the reranker cannot be worse than
the head it starts from.

\textbf{Guard metrics.} The practical question behind Task A is the
\emph{levels} of the elements in a known alloy family, so the headline metric
is present-element WAPE (weighted absolute percentage error over elements
actually present). WAPE alone can be gamed:
a head that predicts every element as present scores deceptively well. Every
composition row therefore also reports all-element WAPE and the macro
false-positive rate.

\textbf{Task B: choosing among the settings that were run.} Temperature and
speed of extrusion are predicted together, and again three heads are compared.
\textbf{Gradient-boosted decision trees} (GBDT) predict each target directly,
with speed modelled in log space \cite{prokhorenkova2018catboost}. A
\textbf{Gaussian process} (GP) does the same on a compressed, condition-pooled
descriptor, with a separate lengthscale per input direction so that
uninformative directions can be discounted; that per-direction weighting is
called automatic relevance determination, so this head is labelled ARD GP
below \cite{rasmussen2006gaussian, duan1983smearing}. The
\textbf{joint-grid GP} uses the same evidence but may only answer with a
setting that was actually run. It places a correlated GP over temperature and
log speed jointly, and instead of reading a value off the posterior mean it
returns the observed training pair that minimises the expected relative error
under that posterior \cite{eikema2022sampling}. Speed estimates from the other
heads are likewise snapped to the nearest of the eleven observed speeds, since
a value between two press settings is never a possible answer
(Appendix~\ref{app:heads}).

\section{Results}
\label{sec:results}

\begin{figure}[t]
  \centering
  \includegraphics[width=0.99\linewidth]{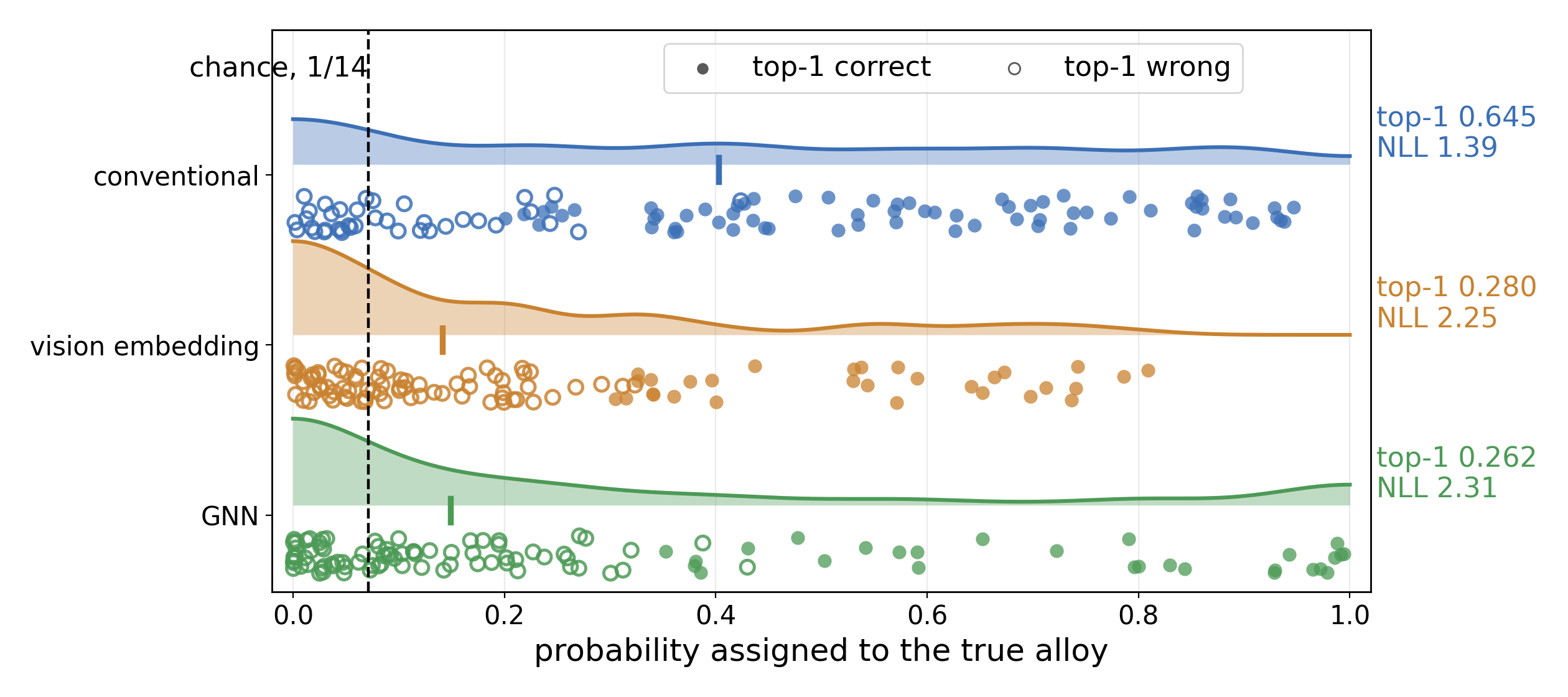}
  \caption{Task A confidence: the probability the reranker assigns to the true
  alloy, pooled over all 107 held-out conditions. Further to the right is
  better, so the conventional row is the best-performing description. Density
  above each line, one dot per condition below (filled = top-1 correct, open =
  wrong), tick = median, dashed = uniform guess ($1/14$). The learned branches
  put little probability on the true alloy for most conditions.}
  \label{fig:confidence}
\end{figure}

\textbf{Task A: composition.} Table~\ref{tab:composition} collects the
5-fold results. The conventional reranker wins outright: 13.6\% WAPE against
16.8\% for the fusion it starts from and 30.0\% for retrieval, with the best
top-1 (0.646)
the fewest false positives (1.1\%), and the sharpest posterior (class NLL
1.38). On the learned branches the selection keeps the fusion unblended on 8
of 10 folds, because the second classifier adds nothing there.
Figure~\ref{fig:confidence} shows what sits underneath: the conventional head
is confidently right on a large share of conditions, while both learned
embeddings assign almost no probability to the truth for many of the alloys,
the mechanism behind their NLL near 2.3. The conventional reranker leads on
every fold (Appendix~\ref{app:folds}). Within one composition step all
branches reach 0.68 to 0.79: each recognises the alloy family, and the gap
lies in resolving its Gd and Mn levels (Appendix~\ref{app:comp_diag}).

The two baselines without microstructure put these numbers in context.
Always predicting the most frequent training alloy gives a top-1 of 0.17, and
a classifier that sees only the known process reaches 0.13, so the process
alone does not reveal the alloy. The conventional reranker is almost four
times the majority rate. The learned branches are about 1.6 times it, and
their NLL (2.23 to 2.30) is only slightly better than that of the class prior
(2.54).

\begin{table}[t]
  \caption{Task A, composition given known process. 5-fold cross-validation,
  mean $\pm$ std over held-out condition groups. WAPE = present-element WAPE;
  FP = false-positive rate; top-1 = alloy accuracy; NLL = class
  log-likelihood. \emph{none}: baselines without microstructure. Best per
  column in bold.}
  \label{tab:composition}
  \centering\small
  \begin{tabularx}{\linewidth}{@{}llXXXX@{}}
    \toprule
    repr. & head & WAPE (\%) & FP (\%) & top-1 & NLL \\
    \midrule
    \multirow{2}{*}{none}
      & most frequent alloy & $71.7 \pm 3.1$ & 75.0 & $0.169 \pm 0.031$ & 2.54 \\
      & known process only & $75.7 \pm 10.8$ & 8.1 & $0.132 \pm 0.063$ & 3.49 \\
    \midrule
    \multirow{3}{*}{conv.}
      & kNN retrieval & $30.0 \pm 7.6$ & 10.3 & $0.496 \pm 0.110$ & 5.92 \\
      & balanced fusion & $16.8 \pm 5.8$ & 1.3 & $0.612 \pm 0.078$ & 1.49 \\
      & OOF-constrained reranker & $\mathbf{13.6 \pm 4.1}$ & \textbf{1.1} & $\mathbf{0.646 \pm 0.051}$ & \textbf{1.38} \\
    \midrule
    \multirow{2}{*}{vision}
      & balanced fusion & $44.7 \pm 18.0$ & 6.5 & $0.276 \pm 0.104$ & 2.25 \\
      & OOF-constrained reranker & $41.6 \pm 18.9$ & 5.9 & $0.287 \pm 0.111$ & 2.23 \\
    \midrule
    \multirow{2}{*}{gnn}
      & balanced fusion & $44.3 \pm 8.0$ & 7.2 & $0.266 \pm 0.071$ & 2.34 \\
      & OOF-constrained reranker & $44.2 \pm 8.0$ & 7.2 & $0.266 \pm 0.071$ & 2.30 \\
    \bottomrule
  \end{tabularx}
\end{table}

\textbf{Task B: process.} Table~\ref{tab:process} shows the mirror task, and
the confound of Section~\ref{sec:data} shows up in its first block. A
classifier that sees only the known composition already names the exact press
speed on 22.5\% of held-out conditions, because each alloy family was
extruded on its own sub-grid; the joint-grid GP reaches only 24 to 29\%. What the
microstructure adds is a smaller error when the guess is not exact. Against
the composition-only baseline, the joint-grid GP roughly halves the
temperature MAE (69.7 to 35-38\,°C), lowers the log speed error from 0.75 to
0.42-0.45, and nearly doubles the share of conditions within one speed
setting (26\% to 46-52\%). It wins every velocity metric among the heads and
gives the best temperature on conventional and vision. The ARD GP collapses on
velocity (MAPE 95-120\%, log MAE 0.63-0.75, worse than predicting the training
mean speed on two of three branches), so the gain comes from decoding on the
observed grid, not from the GP itself. A plain classifier over the observed
settings does not reproduce it either (Appendix~\ref{app:cls_ablation}): it
names the exact speed about as often (23-31\%) but lands within one setting on
only 29-36\%, because it treats neighbouring settings as unrelated. The gain
therefore comes from choosing among the settings that exist while keeping the
distances between them. For temperature, both heads track the diagonal but
saturate above $\approx 400$\degC\ (Figure~\ref{fig:temp_parity} in
Appendix~\ref{app:proc_diag}).

\begin{figure}[t]
  \centering
  \includegraphics[width=0.99\linewidth]{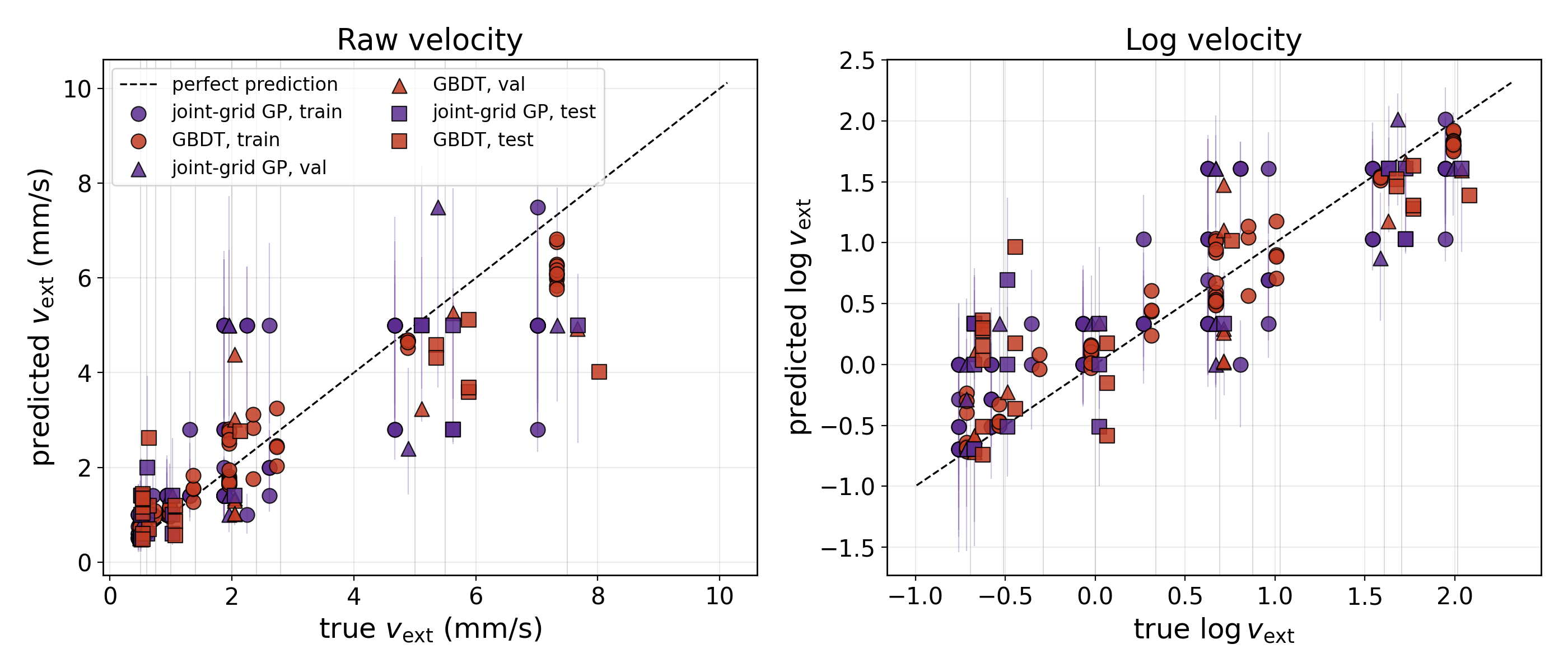}
  \caption{Task B velocity on the GNN branch (\texttt{cv\_fold0}), predicted
  against true $\vext$, raw (left) and log (right). Purple = joint-grid GP,
  red = GBDT, grey = observed press settings, dashed = perfect prediction.
  Both heads are tight at low speed and undershoot the fast conditions.}
  \label{fig:vel_parity}
\end{figure}

\textbf{Velocity is a log-scale quantity.} Speeds span 0.5 to 7.5\,mm/s over
eleven settings, so we also report the error in log space, where a fixed
error is a fixed factor everywhere on the range (Appendix~\ref{app:metrics});
the joint-grid GP's log MAE of 0.42-0.45 is a typical factor of about 1.55.
Figure~\ref{fig:vel_parity} shows that the heads are tight at low speed and
that the spread is asymmetric: they rarely overshoot a slow condition but
often undershoot a fast one.

\begin{figure}[t]
  \centering
  \includegraphics[width=0.99\linewidth]{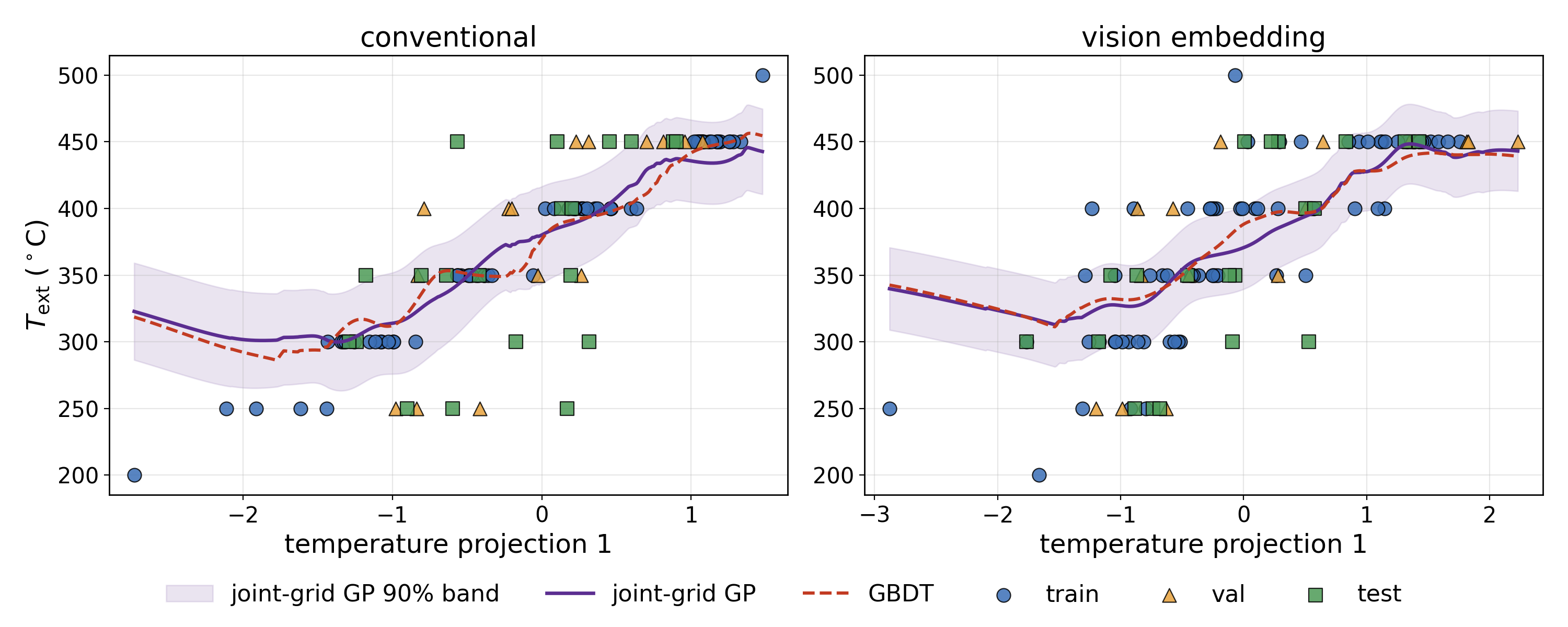}
  \caption{Temperature response along a supervised 1-D projection of the head
  input (selected on validation), one panel per branch. Points = conditions;
  purple curve and band = smoothed joint-grid GP mean and 90\% interval; red
  dashed = GBDT. On every branch the response flattens at both ends, so the
  contraction is already in the descriptor.}
  \label{fig:lda_temperature}
\end{figure}

\textbf{The residual error is tail shrinkage.} Even the best head does not
reach the slowest and fastest settings. On the conventional branch the 26 fastest
conditions (true $\vext \ge 5$\,mm/s, mean 6.3) are predicted at a mean of
4.4\,mm/s, and the 46 slowest ($\le 1$\,mm/s, mean 0.7) at 1.0\,mm/s. With
about two fast examples per alloy family there is little signal to learn the
extremes from, and modelling speed in log space or snapping it to the observed
settings does not recover them. This is a general failure mode of sparse
experimental data: when a target is sampled at a few well-separated settings,
the estimate concentrates on the dense interior and smooths the tails away.
Figure~\ref{fig:lda_temperature} traces this along a 1-D supervised
projection of the head input, and Appendix~\ref{app:proc_diag} shows the same
for velocity: the mean response flattens at
both ends of the range while the held-out points follow the same trend, so
the contraction is already in the descriptor. The remedy is more information,
more extreme-setting conditions or features sensitive to the speed-controlled
recrystallization signature, rather than a different head.

\begin{table}[t]
  \caption{Task B, process given known composition. 5-fold cross-validation,
  mean over held-out condition groups (std in
  Table~\ref{tab:app_process}). MAE in °C ($\Text$) or mm/s ($\vext$); log MAE
  is the velocity error in log space. Hit rate: for $\vext$ the fraction
  recovered at the exact press setting (uniform guess 9\%, most frequent
  training pair 14\%); for $\Text$ the fraction within $\pm 25$°C.
  \emph{none}: a classifier over the observed $(\Text,\vext)$ pairs that sees
  only the known composition. Best per representation block in bold.}
  \label{tab:process}
  \centering\small
  \begin{tabularx}{\linewidth}{@{}lllXXXX@{}}
    \toprule
    repr. & head & target & MAE & MAPE (\%) & log MAE & hit rate (\%) \\
    \midrule
    \multirow{2}{*}{none}
      & composition only & $\Text$ & 69.7 & 19.2 & -- & 29 \\
      & composition only & $\vext$ & 1.71 & 94.9 & 0.748 & 23 \\
    \midrule
    \multirow{6}{*}{conv.}
      & GBDT & $\Text$ & 41.6 & 11.9 & -- & 36 \\
      & ARD GP & $\Text$ & 52.1 & 15.2 & -- & 25 \\
      & joint-grid GP & $\Text$ & \textbf{38.1} & \textbf{10.9} & -- & \textbf{41} \\
      \cmidrule(l){2-7}
      & GBDT & $\vext$ & 1.34 & 64.5 & 0.580 & 11 \\
      & ARD GP & $\vext$ & 1.71 & 104.7 & 0.751 & 15 \\
      & joint-grid GP & $\vext$ & \textbf{0.98} & \textbf{48.7} & \textbf{0.420} & \textbf{24} \\
    \midrule
    \multirow{2}{*}{vision}
      & joint-grid GP & $\Text$ & \textbf{35.2} & \textbf{10.5} & -- & \textbf{46} \\
      & joint-grid GP & $\vext$ & \textbf{1.10} & \textbf{49.6} & \textbf{0.450} & \textbf{24} \\
    \midrule
    \multirow{2}{*}{gnn}
      & GBDT & $\Text$ & 36.5 & 10.7 & -- & 48 \\
      & joint-grid GP & $\vext$ & \textbf{1.01} & \textbf{51.4} & \textbf{0.443} & \textbf{29} \\
    \bottomrule
  \end{tabularx}
\end{table}

\textbf{The guard metrics matter.} While developing the heads, we also tried a
hurdle model and a metric-learning head. Both reached 17.0\% WAPE on one
split, close to the reranker, while predicting 14\% false-positive elements
and 67\% all-element WAPE. On WAPE alone they would have been our headline;
the guards expose them. When most elements are absent, a head can lower its
error by inventing elements, and only the all-element error and the
false-positive rate catch it. Per-fold tables and interval calibration are in
Appendix~\ref{app:results}, and code, stored per-fold predictions and the
scripts that rebuild every table are at
\url{https://github.com/mahishguru/structure-process}.

\section{Conclusion}
\label{sec:conclusion}

Given only an optimized microstructure and its texture, which alloy
composition and what process parameters made it? On a real extrusion campaign
of about a hundred conditions, the answer was limited less by the size of the
representation or the freedom of the regressor than by whether the descriptor
and the form of the predicted answer matched the set of recipes that actually
exist. A recipe is not a continuous quantity: alloys are cast from a short
list of compositions and a press is run at a short list of settings. Prediction
heads that choose from this catalogue while keeping the distances within it
improved on every descriptor we tried. Heads that ignored the catalogue, or
kept it but dropped the distances, did not.

The conventional statistics, measured on the micrograph itself, were easier
for every head to use than embeddings learned for image reconstruction. This
compares descriptor pipelines as practitioners use them, not representation
size (see the limitations below).
Whether larger or retrained embeddings close the gap on
larger campaigns remains open. In practice, optical microscopy and X-ray pole
figures already suffice to
walk the recipe leg. In a self-driving laboratory this makes the recipe leg a
free check on measurements already being collected \cite{low2025selfdriving},
provided the model is reported with the metrics that would expose its failure
modes.

\textbf{Limitations.} The composition task is closed-set: the head picks one of
the 14 known alloys, so unseen compositions are out of reach. The branches do
not see the same input (Table~\ref{tab:inputs}), several heads compress it to
8 to 128 dimensions, and both learned encoders are frozen, so the comparison
does not isolate representation size. The vision input is coloured relative
to a per-alloy mean orientation, so its Task A numbers may be optimistic.
Composition and process are confounded by design, as the composition-only
baseline shows. Finally, 107 conditions from one alloy system and process are
few; larger campaigns with more alloys are the next step.

\begin{ack}
We thank the anonymous reviewers and the area chair for comments that improved
the evaluation of this work.

\textbf{Funding.} We acknowledge Helmholtz-Zentrum Hereon for providing
computational infrastructure and institutional support. Additional support was
provided by the Joachim Herz Stiftung through an Add-on Fellowship.

\textbf{Competing interests.} The authors declare no competing interests.
\end{ack}

\bibliographystyle{unsrtnat}
\bibliography{references}

@article{Kambayashi31122025,
author = {Keita Kambayashi and Ikumu Watanabe},
title = {Computational microstructure design for mechanical property optimization: a review},
journal = {Science and Technology of Advanced Materials: Methods},
volume = {5},
number = {1},
pages = {2581359},
year = {2025},
publisher = {Taylor \& Francis},
doi = {10.1080/27660400.2025.2581359},


URL = { 
    
        https://doi.org/10.1080/27660400.2025.2581359
    
    

},
eprint = { 
    
        https://doi.org/10.1080/27660400.2025.2581359
    
    

}
}

@article{Lin2026,
author = {Lin, Weihao and Jin, Liuchao and Jiang, Zhifei and Cai, Mingjing and Lai, Zhihui and Yurchenko, Daniil and Yan, Bo and Zhou, Shengxi and Novoselov, Kostya S. and Liao, Wei-Hsin and Fang, Shitong},
title = {Machine Learning-Based Inverse Design for Functional Materials: Methods, Challenges, and Engineering Applications},
journal = {Advanced Functional Materials},
volume = {36},
number = {40},
pages = {e75070},
doi = {https://doi.org/10.1002/adfm.75070},
url = {https://advanced.onlinelibrary.wiley.com/doi/abs/10.1002/adfm.75070},
eprint = {https://advanced.onlinelibrary.wiley.com/doi/pdf/10.1002/adfm.75070},
year = {2026}
}

@article{Wu2021,
  author    = {Wu, Jun and Sigmund, Ole and Groen, Jeroen P.},
  title     = {Topology optimization of multi-scale structures: a review},
  journal   = {Structural and Multidisciplinary Optimization},
  volume    = {63},
  number    = {3},
  pages     = {1455--1480},
  year      = {2021},
  month     = {mar},
  doi       = {10.1007/s00158-021-02881-8},
  url       = {https://doi.org/10.1007/s00158-021-02881-8},
  issn      = {1615-1488}
}

@article{INDURKAR2020102762,
title = {Predicting textural variability effects in the anisotropic plasticity and stability of hexagonal metals: Application to magnesium and its alloys},
journal = {International Journal of Plasticity},
volume = {132},
pages = {102762},
year = {2020},
issn = {0749-6419},
doi = {https://doi.org/10.1016/j.ijplas.2020.102762},
url = {https://www.sciencedirect.com/science/article/pii/S0749641919309325},
author = {Padmeya P. Indurkar and Shahmeer Baweja and Robert Perez and Shailendra P. Joshi}
}

@article{PAULSON2017428,
title = {Reduced-order structure-property linkages for polycrystalline microstructures based on 2-point statistics},
journal = {Acta Materialia},
volume = {129},
pages = {428-438},
year = {2017},
issn = {1359-6454},
doi = {https://doi.org/10.1016/j.actamat.2017.03.009},
url = {https://www.sciencedirect.com/science/article/pii/S135964541730188X},
author = {Noah H. Paulson and Matthew W. Priddy and David L. McDowell and Surya R. Kalidindi}
}

@article{KUSAMPUDI2023103776,
title = {Inverse design of dual-phase steel microstructures using generative machine learning model and Bayesian optimization},
journal = {International Journal of Plasticity},
volume = {171},
pages = {103776},
year = {2023},
issn = {0749-6419},
doi = {https://doi.org/10.1016/j.ijplas.2023.103776},
url = {https://www.sciencedirect.com/science/article/pii/S0749641923002607},
author = {Navyanth Kusampudi and Martin Diehl}
}

@article{Guru2025,
  author  = {Guru, Mahish K. and Bohlen, Jan and Aydin, Roland C. and Ben Khalifa, Noomane},
  title   = {Machine learning pipeline for structure-property modeling in Mg-alloys using microstructure and texture descriptors},
  journal = {Acta Materialia},
  year    = {2025}
}

@article{Tom2024,
    author = {Tom, Gary and Schmid, Stefan P. and Baird, Sterling G. and Cao, Yang and Darvish, Kourosh and Hao, Han and Lo, Stanley and Pablo-Garc{\'i}a, Sergio and Rajaonson, Ella M. and Skreta, Marta and Yoshikawa, Naruki and Corapi, Samantha and Akkoc, Gun Deniz and Strieth-Kalthoff, Felix and Seifrid, Martin and Aspuru-Guzik, Al{\'a}n},
    title = {Self-Driving
Laboratories for Chemistry and Materials
Science},
    journal = {Chemical Reviews},
    volume = {124},
    number = {16},
    pages = {9633-9732},
    year = {2024},
    month = {08},
    issn = {0009-2665},
    doi = {10.1021/acs.chemrev.4c00055},
    url = {https://doi.org/10.1021/acs.chemrev.4c00055},
    eprint = {https://pubs.acs.org/chreay/article-pdf/124/16/9633/1507454/cr4c00055.pdf},
}

@article{Lee2026,
    author = {Lee, Heeseung and Yoo, Hyuk Jun and Jang, Hye Su and Park, Byeongho and Park, Yang Jeong and Han, Sang Soo},
    title = {Toward self-driving laboratory 2.0 for chemistry and materials discovery},
    journal = {Materials Horizons},
    volume = {13},
    number = {10},
    pages = {4712-4739},
    year = {2026},
    month = {05},
    issn = {2051-6347},
    doi = {10.1039/d5mh01984b},
    url = {https://doi.org/10.1039/d5mh01984b},
    eprint = {https://pubs.rsc.org/mh/article-pdf/13/10/4712/12525298/d5mh01984b.pdf},
}

@article{Adesiji,
    author = {Adesiji, Adedire D. and Wang, Jiashuo and Kuo, Cheng-Shu and Brown, Keith A.},
    title = {Benchmarking self-driving labs},
    journal = {Digital Discovery},
    volume = {5},
    number = {1},
    pages = {14-27},
    year = {2026},
    month = {01},
    issn = {2635-098X},
    doi = {10.1039/d5dd00337g},
    url = {https://doi.org/10.1039/d5dd00337g},
    eprint = {https://pubs.rsc.org/dd/article-pdf/5/1/14/10334851/d5dd00337g.pdf},
}

@article{Kalinin2021,
    author = {Kalinin, Sergei V. and Ziatdinov, Maxim and Hinkle, Jacob and Jesse, Stephen and Ghosh, Ayana and Kelley, Kyle P. and Lupini, Andrew R. and Sumpter, Bobby G. and Vasudevan, Rama K.},
    title = {Automated
and Autonomous Experiments in Electron and
Scanning Probe Microscopy},
    journal = {ACS Nano},
    volume = {15},
    number = {8},
    pages = {12604-12627},
    year = {2021},
    month = {07},
    issn = {1936-0851},
    doi = {10.1021/acsnano.1c02104},
    url = {https://doi.org/10.1021/acsnano.1c02104},
    eprint = {https://pubs.acs.org/ancac3/article-pdf/15/8/12604/16501312/nn1c02104.pdf},
}

@article{Kalinin2022DeepLF,
  title={Deep learning for electron and scanning probe microscopy: From materials design to atomic fabrication},
  author={Sergei V. Kalinin and Maxim A. Ziatdinov and Steven R. Spurgeon and Colin Ophus and Eric A. Stach and Toma Susi and Josh Agar and John N. Randall},
  journal={MRS Bulletin},
  year={2022},
  volume={47},
  pages={931-939},
  url={https://api.semanticscholar.org/CorpusID:253326381}
}

@ARTICLE{Nienaber2019,
    
AUTHOR={Nienaber, Maria  and Kainer, Karl Ulrich  and Letzig, Dietmar  and Bohlen, Jan },
           
TITLE={Processing Effects on the Formability of Extruded Flat Products of Magnesium Alloys},
          
JOURNAL={Frontiers in Materials},
          
VOLUME={Volume 6 - 2019},
  
YEAR={2019},
  
URL={https://www.frontiersin.org/journals/materials/articles/10.3389/fmats.2019.00253},
  
DOI={10.3389/fmats.2019.00253},
  
ISSN={2296-8016}}

@article{Kalidindi-MKS-2015,
   author = "Kalidindi, Surya R. and De Graef, Marc",
   title = "Materials Data Science: Current Status and Future Outlook", 
   journal= "Annual Review of Materials Research",
   year = "2015",
   volume = "45",
   number = "Volume 45, 2015",
   pages = "171-193",
   doi = "https://doi.org/10.1146/annurev-matsci-070214-020844",
   url = "https://www.annualreviews.org/content/journals/10.1146/annurev-matsci-070214-020844",
   publisher = "Annual Reviews",
   issn = "1545-4118",
   type = "Journal Article",
  }

@article{groeber2014,
author = {Groeber, Michael and Jackson, Michael},
year = {2014},
month = {02},
pages = {5},
title = {DREAM.3D: A Digital Representation Environment for the Analysis of Microstructure in 3D},
volume = {3},
journal = {Integrating Materials and Manufacturing Innovation},
doi = {10.1186/2193-9772-3-5}
}

@article{markley2007,
author = {Markley, Landis and Cheng, Yang and Crassidis, John and Oshman, Yaakov},
year = {2007},
month = {07},
pages = {1193-1196},
title = {Averaging Quaternions},
volume = {30},
journal = {Journal of Guidance, Control, and Dynamics},
doi = {10.2514/1.28949}
}

@inproceedings{radford2021clip,
  title     = {Learning Transferable Visual Models From Natural Language Supervision},
  author    = {Alec Radford and Jong Wook Kim and Chris Hallacy and Aditya Ramesh and Gabriel Goh and Sandhini Agarwal and Girish Sastry and Amanda Askell and Pamela Mishkin and Jack Clark and Gretchen Krueger and Ilya Sutskever},
  booktitle = {Proceedings of the 38th International Conference on Machine Learning},
  editor    = {Marina Meila and Tong Zhang},
  series    = {Proceedings of Machine Learning Research},
  volume    = {139},
  pages     = {8748--8763},
  year      = {2021},
  publisher = {PMLR},
  url       = {https://proceedings.mlr.press/v139/radford21a.html}
}

@article{esser2024sd3,
  title   = {Scaling Rectified Flow Transformers for High-Resolution Image Synthesis},
  author  = {Patrick Esser and Sumith Kulal and Andreas Blattmann and Rahim Entezari and Jonas M{\"u}ller and Harry Saini and Yam Levi and Dominik Lorenz and Axel Sauer and Frederic Boesel and Dustin Podell and Tim Dockhorn and Zion English and Kyle Lacey and Alex Goodwin and Yannik Marek and Robin Rombach},
  journal = {arXiv preprint arXiv:2403.03206},
  year    = {2024},
  url     = {https://arxiv.org/abs/2403.03206}
}

@inproceedings{peebles2023dit,
  author    = {William Peebles and Saining Xie},
  title     = {Scalable Diffusion Models with Transformers},
  booktitle = {Proceedings of the IEEE/CVF International Conference on Computer Vision (ICCV)},
  month     = {October},
  year      = {2023},
  pages     = {4195--4205},
  url       = {https://openaccess.thecvf.com/content/ICCV2023/html/Peebles_Scalable_Diffusion_Models_with_Transformers_ICCV_2023_paper.html}
}

@inproceedings{lipman2023flowmatching,
  title     = {Flow Matching for Generative Modeling},
  author    = {Yaron Lipman and Ricky T. Q. Chen and Heli Ben-Hamu and Maximilian Nickel and Matt Le},
  booktitle = {Proceedings of the International Conference on Learning Representations},
  year      = {2023},
  url       = {https://openreview.net/forum?id=PqvMRDCJT9t}
}

@phdthesis{nienaber_einfluss_der_2023,
  author = {Nienaber, Maria},
  title  = {Einfluss der Prozessparameter auf die Gefüge- und Textureinstellung sowie die mechanischen Eigenschaften stranggepresster Magnesiumbleche},
  school = {Helmholtz-Zentrum Geesthacht / Technische Universität Hamburg},
  year   = {2023}
}

@article{CANOCASTILLO2020139527,
  author  = {Cano-Castillo, José Luis and others},
  title   = {Microstructure and texture development of Zn-bearing magnesium alloys during extrusion},
  journal = {Materials Science and Engineering: A},
  volume  = {795},
  pages   = {139527},
  year    = {2020},
  doi     = {10.1016/j.msea.2020.139527}
}

@article{KurzG,
  author = {Kurz, Gerald and others},
  title  = {Tailoring the mechanical properties of the aluminum-free magnesium alloy ME21 by extrusion},
  journal = {Materials Science and Engineering: A},
  year   = {2015}
}

@article{Harmuth,
  author = {Harmuth, Julia and others},
  title  = {Effect of extrusion on the microstructure, texture and mechanical properties of Mg-Gd alloys},
  journal = {Materials Science and Engineering: A},
  year   = {2019}
}

@article{cryst12081036,
  author  = {others},
  title   = {Microstructure and texture of extruded Mg-Gd-Mn alloys},
  journal = {Crystals},
  volume  = {12},
  number  = {8},
  pages   = {1036},
  year    = {2022},
  doi     = {10.3390/cryst12081036}
}

@article{tenenbaum2000isomap,
  author  = {Tenenbaum, Joshua B. and de Silva, Vin and Langford, John C.},
  title   = {A Global Geometric Framework for Nonlinear Dimensionality Reduction},
  journal = {Science},
  volume  = {290},
  number  = {5500},
  pages   = {2319--2323},
  year    = {2000},
  doi     = {10.1126/science.290.5500.2319}
}

@inproceedings{gatys2015texture,
  author    = {Gatys, Leon A. and Ecker, Alexander S. and Bethge, Matthias},
  title     = {Texture Synthesis Using Convolutional Neural Networks},
  booktitle = {Advances in Neural Information Processing Systems 28},
  pages     = {262--270},
  year      = {2015}
}

@inproceedings{gatys2016style,
  author    = {Gatys, Leon A. and Ecker, Alexander S. and Bethge, Matthias},
  title     = {Image Style Transfer Using Convolutional Neural Networks},
  booktitle = {Proc. IEEE Conf. Computer Vision and Pattern Recognition (CVPR)},
  pages     = {2414--2423},
  year      = {2016}
}

@inproceedings{dosovitskiy2021image,
  title     = {An Image is Worth 16x16 Words: Transformers for Image Recognition at Scale},
  author    = {Dosovitskiy, Alexey and Beyer, Lucas and Kolesnikov, Alexander and
               Weissenborn, Dirk and Zhai, Xiaohua and Unterthiner, Thomas and
               Dehghani, Mostafa and Minderer, Matthias and Heigold, Georg and
               Gelly, Sylvain and Uszkoreit, Jakob and Houlsby, Neil},
  booktitle = {International Conference on Learning Representations},
  year      = {2021},
  url       = {https://arxiv.org/abs/2010.11929}
}

@inproceedings{brody2022how,
  title     = {How Attentive are Graph Attention Networks?},
  author    = {Brody, Shaked and Alon, Uri and Yahav, Eran},
  booktitle = {International Conference on Learning Representations},
  year      = {2022},
  url       = {https://arxiv.org/abs/2105.14491}
}

@inproceedings{velickovic2018graph,
  title     = {Graph Attention Networks},
  author    = {Veli{\v{c}}kovi{\'c}, Petar and Cucurull, Guillem and Casanova, Arantxa and
               Romero, Adriana and Lio, Pietro and Bengio, Yoshua},
  booktitle = {International Conference on Learning Representations},
  year      = {2018},
  url       = {https://arxiv.org/abs/1710.10903}
}

@inproceedings{vaswani2017attention,
  title     = {Attention Is All You Need},
  author    = {Vaswani, Ashish and Shazeer, Noam and Parmar, Niki and
               Uszkoreit, Jakob and Jones, Llion and Gomez, Aidan N. and
               Kaiser, {\L}ukasz and Polosukhin, Illia},
  booktitle = {Advances in Neural Information Processing Systems},
  volume    = {30},
  year      = {2017},
  url       = {https://arxiv.org/abs/1706.03762}
}

@article{cover1967nearest,
  author  = {Cover, Thomas and Hart, Peter},
  title   = {Nearest Neighbor Pattern Classification},
  journal = {IEEE Transactions on Information Theory},
  volume  = {13},
  number  = {1},
  pages   = {21--27},
  year    = {1967},
  doi     = {10.1109/TIT.1967.1053964},
  url     = {https://ieeexplore.ieee.org/document/1053964}
}

@inproceedings{prokhorenkova2018catboost,
  author    = {Prokhorenkova, Liudmila and Gusev, Gleb and Vorobev, Aleksandr and
               Dorogush, Anna Veronika and Gulin, Andrey},
  title     = {CatBoost: Unbiased Boosting with Categorical Features},
  booktitle = {Advances in Neural Information Processing Systems},
  volume    = {31},
  year      = {2018},
  url       = {https://arxiv.org/abs/1706.09516}
}

@article{friedman2001greedy,
  author  = {Friedman, Jerome H.},
  title   = {Greedy Function Approximation: A Gradient Boosting Machine},
  journal = {The Annals of Statistics},
  volume  = {29},
  number  = {5},
  pages   = {1189--1232},
  year    = {2001},
  doi     = {10.1214/aos/1013203451},
  url     = {https://doi.org/10.1214/aos/1013203451}
}

@inproceedings{chen2016xgboost,
  author    = {Chen, Tianqi and Guestrin, Carlos},
  title     = {XGBoost: A Scalable Tree Boosting System},
  booktitle = {Proceedings of the 22nd ACM SIGKDD International Conference on
               Knowledge Discovery and Data Mining},
  pages     = {785--794},
  year      = {2016},
  doi       = {10.1145/2939672.2939785},
  url       = {https://arxiv.org/abs/1603.02754}
}

@book{hastie2009elements,
  author    = {Hastie, Trevor and Tibshirani, Robert and Friedman, Jerome},
  title     = {The Elements of Statistical Learning},
  edition   = {2},
  publisher = {Springer},
  year      = {2009},
  doi       = {10.1007/978-0-387-84858-7},
  url       = {https://doi.org/10.1007/978-0-387-84858-7}
}

@article{wolpert1992stacked,
  author  = {Wolpert, David H.},
  title   = {Stacked Generalization},
  journal = {Neural Networks},
  volume  = {5},
  number  = {2},
  pages   = {241--259},
  year    = {1992},
  doi     = {10.1016/S0893-6080(05)80023-1},
  url     = {https://doi.org/10.1016/S0893-6080(05)80023-1}
}

@article{varma2006bias,
  author  = {Varma, Sudhir and Simon, Richard},
  title   = {Bias in Error Estimation When Using Cross-Validation for Model Selection},
  journal = {BMC Bioinformatics},
  volume  = {7},
  pages   = {91},
  year    = {2006},
  doi     = {10.1186/1471-2105-7-91},
  url     = {https://doi.org/10.1186/1471-2105-7-91}
}

@article{badaro2023transformers,
  author  = {Badaro, Gilbert and Saeed, Mohammed and Papotti, Paolo},
  title   = {Transformers for Tabular Data Representation: A Survey of Models and Applications},
  journal = {Transactions of the Association for Computational Linguistics},
  volume  = {11},
  pages   = {227--249},
  year    = {2023},
  publisher = {MIT Press},
  doi     = {10.1162/tacl_a_00544},
  url     = {https://aclanthology.org/2023.tacl-1.14/}
}

@book{rasmussen2006gaussian,
  author    = {Rasmussen, Carl Edward and Williams, Christopher K. I.},
  title     = {Gaussian Processes for Machine Learning},
  publisher = {MIT Press},
  address   = {Cambridge, MA},
  year      = {2006},
  url       = {https://gaussianprocess.org/gpml/}
}

@article{duan1983smearing,
  author  = {Duan, Naihua},
  title   = {Smearing Estimate: A Nonparametric Retransformation Method},
  journal = {Journal of the American Statistical Association},
  volume  = {78},
  number  = {383},
  pages   = {605--610},
  year    = {1983},
  doi     = {10.1080/01621459.1983.10478017},
  url     = {https://doi.org/10.1080/01621459.1983.10478017}
}

@inproceedings{eikema2022sampling,
  author    = {Eikema, Bryan and Aziz, Wilker},
  title     = {Sampling-Based Approximations to Minimum Bayes Risk Decoding for Neural Machine Translation},
  booktitle = {Proceedings of the 2022 Conference on Empirical Methods in Natural Language Processing},
  year      = {2022},
  url       = {https://arxiv.org/abs/2108.04718}
}

@article{low2025selfdriving,
  author  = {Low, Andre K. Y. and Cheng, Jayce J. W. and Hippalgaonkar, Kedar and
             Ng, Leonard W. T.},
  title   = {Self-Driving Laboratories: Translating Materials Science from Laboratory to Factory},
  journal = {ACS Omega},
  volume  = {10},
  number  = {28},
  pages   = {29902--29908},
  year    = {2025},
  doi     = {10.1021/acsomega.5c02197},
  url     = {https://doi.org/10.1021/acsomega.5c02197}
}

@inproceedings{gorishniy2021revisiting,
  author    = {Gorishniy, Yury and Rubachev, Ivan and Khrulkov, Valentin and Babenko, Artem},
  title     = {Revisiting Deep Learning Models for Tabular Data},
  booktitle = {Advances in Neural Information Processing Systems},
  volume    = {34},
  year      = {2021},
  url       = {https://arxiv.org/abs/2106.11959}
}

@article{randle2001grain,
  author  = {Randle, Valerie and Davies, Helen and Cross, Ian},
  title   = {Grain boundary misorientation distributions},
  journal = {Current Opinion in Solid State and Materials Science},
  volume  = {5},
  number  = {1},
  pages   = {3--8},
  year    = {2001},
  doi     = {10.1016/S1359-0286(00)00018-8},
  url     = {https://doi.org/10.1016/S1359-0286(00)00018-8}
}

@inproceedings{wang2014learning,
  author    = {Wang, Jiang and Song, Yang and Leung, Thomas and Rosenberg, Chuck
               and Wang, Jingbin and Philbin, James and Chen, Bo and Wu, Ying},
  title     = {Learning Fine-grained Image Similarity with Deep Ranking},
  booktitle = {Proceedings of the IEEE Conference on Computer Vision and Pattern Recognition (CVPR)},
  pages     = {1386--1393},
  year      = {2014},
  doi       = {10.1109/CVPR.2014.180},
  url       = {https://arxiv.org/abs/1404.4661}
}

@misc{guru2026copilotconstrainedphysicsinformedlatent,
  title         = {Co-PiLOT: Constrained Physics-Informed Latent Optimization for Target-Driven Inverse Design},
  author        = {Mahish K. Guru and Mayank Nagar and Ayush Vyas and Jan Bohlen and Roland Aydin and Noomane Ben Khalifa},
  year          = {2026},
  eprint        = {2609.37875},
  archivePrefix = {arXiv},
  primaryClass  = {cs.AI},
  url           = {https://arxiv.org/abs/2609.37875}
}

@article{bostanabad2018computational,
  author  = {Bostanabad, Ramin and Zhang, Yichi and Li, Xiaolin and Kearney, Tucker and Brinson, L. Catherine and Apley, Daniel W. and Liu, Wing Kam and Chen, Wei},
  title   = {Computational microstructure characterization and reconstruction: Review of the state-of-the-art techniques},
  journal = {Progress in Materials Science},
  volume  = {95},
  pages   = {1--41},
  year    = {2018},
  doi     = {10.1016/j.pmatsci.2018.01.005}
}

@article{popova2017process,
  author  = {Popova, Evdokia and Rodgers, Theron M. and Gong, Xinyi and Cecen, Ahmet and Madison, Jonathan D. and Kalidindi, Surya R.},
  title   = {Process-Structure Linkages Using a Data Science Approach: Application to Simulated Additive Manufacturing Data},
  journal = {Integrating Materials and Manufacturing Innovation},
  volume  = {6},
  number  = {1},
  pages   = {54--68},
  year    = {2017},
  doi     = {10.1007/s40192-017-0088-1}
}

@article{decost2015computer,
  author  = {DeCost, Brian L. and Holm, Elizabeth A.},
  title   = {A computer vision approach for automated analysis and classification of microstructural image data},
  journal = {Computational Materials Science},
  volume  = {110},
  pages   = {126--133},
  year    = {2015},
  doi     = {10.1016/j.commatsci.2015.08.011}
}

@article{decost2017exploring,
  author  = {DeCost, Brian L. and Francis, Toby and Holm, Elizabeth A.},
  title   = {Exploring the microstructure manifold: Image texture representations applied to ultrahigh carbon steel microstructures},
  journal = {Acta Materialia},
  volume  = {133},
  pages   = {30--40},
  year    = {2017},
  doi     = {10.1016/j.actamat.2017.05.014}
}

@article{holm2020overview,
  author  = {Holm, Elizabeth A. and Cohn, Ryan and Gao, Nan and Kitahara, Andrew R. and Matson, Thomas P. and Lei, Bo and Yarasi, Srujana Rao},
  title   = {Overview: Computer Vision and Machine Learning for Microstructural Characterization and Analysis},
  journal = {Metallurgical and Materials Transactions A},
  volume  = {51},
  pages   = {5985--5999},
  year    = {2020},
  doi     = {10.1007/s11661-020-06008-4}
}

@article{stuckner2022microstructure,
  author  = {Stuckner, Joshua and Harder, Bryan and Smith, Timothy M.},
  title   = {Microstructure segmentation with deep learning encoders pre-trained on a large microscopy dataset},
  journal = {npj Computational Materials},
  volume  = {8},
  pages   = {200},
  year    = {2022},
  doi     = {10.1038/s41524-022-00878-5}
}

@article{dai2021graph,
  author  = {Dai, Minyi and Demirel, Mehmet F. and Liang, Yingyu and Hu, Jia-Mian},
  title   = {Graph neural networks for an accurate and interpretable prediction of the properties of polycrystalline materials},
  journal = {npj Computational Materials},
  volume  = {7},
  pages   = {103},
  year    = {2021},
  doi     = {10.1038/s41524-021-00574-w}
}

\appendix

\section{Method details}
\label{app:method}

The appendix has two parts. This first part completes the description of the
method: Appendix~\ref{app:descriptors} specifies the three descriptors of
Section~\ref{sec:repr}, Appendix~\ref{app:heads} the prediction heads of
Section~\ref{sec:heads}, and Appendix~\ref{app:metrics} the metrics and
baselines. Appendix~\ref{app:results} then reports the results
that did not fit into Section~\ref{sec:results}: the per-fold tables behind
every mean quoted in the main text, a classification ablation for Task B, and
the diagnostics that show where the remaining error comes from.

\subsection{Descriptor specifications}
\label{app:descriptors}

This subsection gives the detail behind Section~\ref{sec:repr}. In the
conventional branch the 438 columns are ordered as 3-point spatial
correlations reduced by PCA (columns 0-120), Gram matrices of pretrained CNN
activations reduced by Isomap (120-320), the aspect-ratio distribution
(320-349), the equivalent-diameter distribution (349-378) and generalized
spherical harmonics coefficients of the ODF reduced by Isomap (378-438). Four
further columns carry the known half of the recipe: $\Text$, $\log \vext$ and
a two-column one-hot for extrusion-ratio type.

The two learned branches share one RVE per micrograph. Each RVE is built in
DREAM.3D \cite{groeber2014} from a Gaussian copula fitted to that
micrograph's grain statistics, with a lognormal equivalent diameter and a beta
aspect ratio; micrographs with fewer than 10 usable grains are skipped. Grain
orientations are sampled from the condition's ODF, so all images of a
condition share its texture.

In Co-PiLOT \cite{guru2026copilotconstrainedphysicsinformedlatent} the
ViT-H/14 trunk kept its lower 16 blocks at the CLIP weights and fine-tuned its
upper 16 blocks, and a multi-query attention pooler produces 16 spatial latent
tokens of width 80. Architecture and training of the encoder and the MMDiT
decoder are described there. In this paper the full encoder is frozen and is
used only to extract features; the flattened $16 \times 80$ latent is the
descriptor.

\begin{figure}[H]
  \centering
  \includegraphics[width=0.99\linewidth]{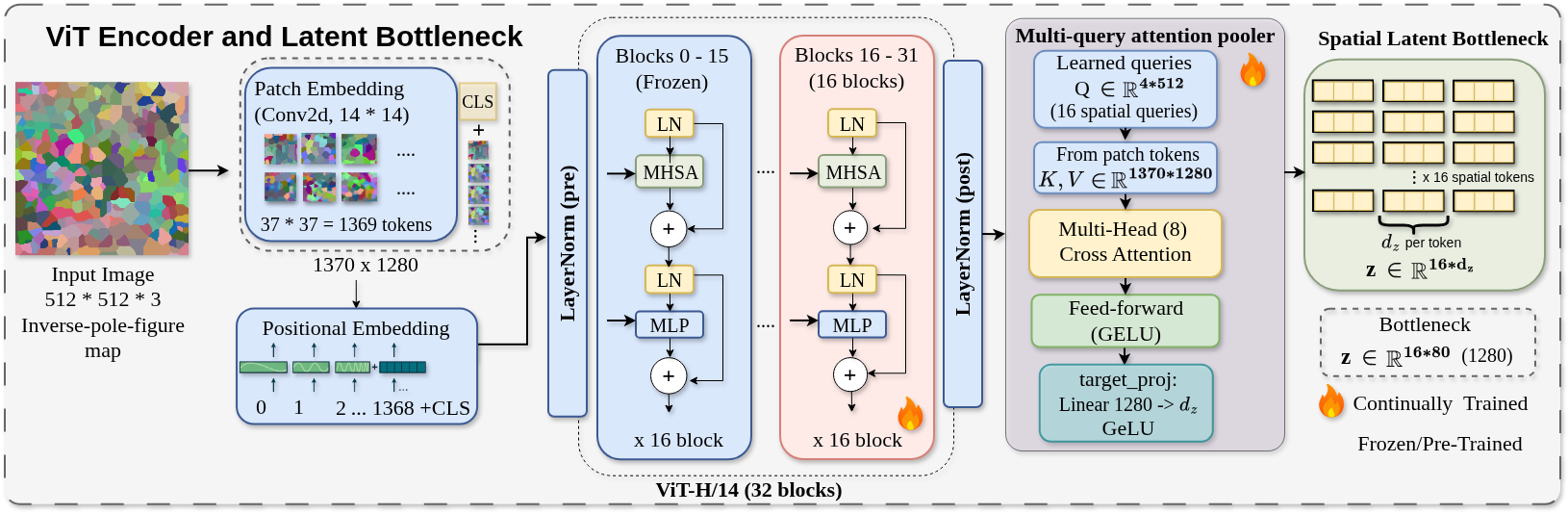}
  \caption{The vision-embedding branch: the frozen Co-PiLOT ViT-H encoder maps
  an orientation-coloured RVE map to the latent from which the MMDiT decoder
  reconstructs it.}
  \label{fig:vision}
\end{figure}

In the grain-graph branch, node features are grain area, aspect ratio,
Bunge-Euler angles, centroid and bounding box, and edge features are
shared-boundary length and misorientation angle. A linear projection to 128
dimensions is followed by four GATv2 blocks (4 attention heads, edge dimension
2, LayerNorm, GELU and residual connections), then attention pooling with 8
learned seed queries. Graphs contain roughly 500 grains per image.

\subsection{Prediction head specifications}
\label{app:heads}

This subsection gives the detail behind Section~\ref{sec:heads}. Throughout,
\emph{inner splits} are condition-stratified splits of a training fold, used
to produce out-of-fold (OOF) predictions on training conditions only; all
preprocessing is refitted inside every inner split. Validation portions select
base hyperparameters, and the held-out test fold selects nothing.

\textbf{Task A, decode.} A head outputs a probability over the 14 alloys
per image. Probabilities are averaged over the images of a condition, the
top-1 alloy is taken, and its nominal wt\% vector over the eight elements is
the predicted composition. Alloy top-1 and the class NLL are read from the
same pooled probability, so every composition metric in
Table~\ref{tab:composition} refers to one prediction per condition.

\textbf{Task A, retrieval.} Descriptors are L2-normalized and compared by
cosine similarity. The $k$ nearest training images vote for their alloy with
softmax weights $\exp(s/\tau)$ over the similarities $s$; $k \in \{1,3,5,7\}$
and $\tau \in \{0.01, 0.05, 0.1, 0.2\}$ are selected on validation. In the
balanced variant each vote is additionally scaled by the inverse image count
of its condition.

\textbf{Task A, condition-balanced fusion.} A 14-way CatBoost classifier
\cite{prokhorenkova2018catboost} is trained on the image-level descriptor
concatenated with the known process columns ($\Text$, $\log \vext$, and a
two-column one-hot for extrusion-ratio type), with sample weights set so that
every condition contributes the same total weight. Its probability is averaged
50:50 with the balanced retrieval vote.

\textbf{Task A, reranker.} The fusion probability $p_{\mathrm{fus}}$ is
blended with a second classifier $p_{\mathrm{aux}}$ as
$p = w\, p_{\mathrm{fus}} + (1-w)\, p_{\mathrm{aux}}$ with
$w \in \{0, 0.25, 0.5, 0.75, 1\}$. On the conventional branch
$p_{\mathrm{aux}}$ is the uniform average of five logistic classifiers, one
per descriptor block: 3-point correlations (columns 0-120), Gram-Isomap
(120-320), aspect ratio (320-349), equivalent diameter (349-378) and
GSH-Isomap (378-438), each fitted on condition-mean features with the known
columns appended, standardized, with $C \in \{0.01, 0.1, 1, 10\}$ selected on
the inner splits. On the vision and gnn branches, which have no block
structure, $p_{\mathrm{aux}}$ is an FT-Transformer classifier
\cite{gorishniy2021revisiting} on the same condition-mean features. When its
input is wider than 512 columns (the vision branch), it is first reduced by
PCA, fitted on the training conditions, to the components that explain 95\%
of the variance, at most 128. The weight $w$ is chosen on OOF predictions of
the training conditions, restricted to candidates whose OOF all-element WAPE
and macro false-positive rate do not exceed those of the fusion. Because
$w = 1$ is always eligible, the reranker never underperforms the fusion by
construction on the selection objective.

\textbf{Task B, trees and Gaussian process.} The tree head is CatBoost with
standardized targets and $\vext$ modelled as $\log \vext$. The ARD GP pools
descriptors to one row per condition, standardizes, reduces to at most 32
principal components, and fits an independent GP per target with kernel
$c \cdot \mathrm{RBF}(\boldsymbol{\ell}) + \mathrm{White}$, where
$\boldsymbol{\ell}$ is a per-dimension lengthscale vector
\cite{rasmussen2006gaussian}. Because $\vext$ is fitted in log space, the
reported point prediction applies the lognormal smearing correction
\cite{duan1983smearing}; the median is retained separately.

\textbf{Task B, joint-grid GP.} The feature map compresses the condition's
descriptor distribution with PCA to 8 components followed by a random Fourier
feature expansion, and the resulting kernel is mixed 50:50 with an RBF kernel
over the known composition and extrusion-ratio columns. A two-task correlated
GP is fitted jointly over $(\Text, \log \vext)$, so evidence about temperature
informs speed and the reverse. At a query the posterior is not read off the
mean. Instead the head scores every $(\Text, \vext)$ pair observed in the
training fold and returns the pair with the smallest expected relative error
under its own posterior, with a small penalty term keeping the choice near the
GP point estimate. Velocity estimates that come from any other head
are snapped onto the observed speed levels, nearest in $\log \vext$, because
the 107 conditions contain only eleven distinct speeds spanning 0.5 to
7.5\,mm/s and are roughly evenly spaced in log space. An optional ordinal classifier
over the speed levels, built as a cumulative sequence of gradient-boosted
binary classifiers, may be blended in; that choice and the choice of which
estimate to snap are made on OOF training predictions under the constraint
that they may not degrade the plain snapped log-space GP.

\subsection{Metrics and baselines}
\label{app:metrics}

All metrics are computed per held-out fold on one prediction per condition
and then averaged over the five folds. Let $c_{ie}$ and $\hat c_{ie}$ be the
true and predicted wt\% of element $e$ in condition $i$, $E = 8$ the number of
elements, and $P_e = \{i : c_{ie} > 0\}$ the conditions that contain $e$.
\begin{align*}
  \mathrm{WAPE} &= \frac{1}{E}\sum_{e}
    \frac{\sum_{i \in P_e} |\hat c_{ie} - c_{ie}|}{\sum_{i \in P_e} c_{ie}},
  &
  \mathrm{WAPE}_{\mathrm{all}} &= \frac{1}{E}\sum_{e}
    \frac{\sum_{i} |\hat c_{ie} - c_{ie}|}{\sum_{i \in P_e} c_{ie}},
  \\
  \mathrm{FP} &= \frac{1}{E}\sum_{e}
    \frac{|\{i \notin P_e : \hat c_{ie} > 0\}|}{|\{i \notin P_e\}|},
  &
  \mathrm{NLL} &= -\frac{1}{N}\sum_{i} \log \bar p_i(a_i).
\end{align*}
Here $\bar p_i$ is the image-level alloy probability averaged over the images
of condition $i$, $a_i$ its true alloy and $N$ the number of held-out
conditions; top-1 is the fraction of conditions with
$\arg\max \bar p_i = a_i$. WAPE divides by the amount of each element that is
present, so elements at 0.2 and at 10 wt\% count equally, and the macro
average keeps the rare elements visible. WAPE is restricted to present
elements because the practical question is the levels within a known family.
$\mathrm{WAPE}_{\mathrm{all}}$ and FP are the guards: they rise when a head
predicts elements that are not there.

For the process, with $y_i$ the true and $\hat y_i$ the predicted $\Text$ or
$\vext$,
\begin{align*}
  \mathrm{MAE} &= \frac{1}{N}\sum_i |\hat y_i - y_i|,
  &
  \mathrm{MAPE} &= \frac{100}{N}\sum_i \frac{|\hat y_i - y_i|}{y_i},
  &
  \mathrm{log\ MAE} &= \frac{1}{N}\sum_i |\log \hat v_i - \log v_i|.
\end{align*}
The log MAE is used because the speeds span a factor of 15, and
$\exp(\mathrm{log\ MAE})$ is the typical multiplicative error. Let
$\ell(v)$ be the index of the observed speed level nearest to $v$ in log
space. The exact-speed rate is the fraction of conditions with
$\ell(\hat v_i) = \ell(v_i)$, the within-one rate the fraction with
$|\ell(\hat v_i) - \ell(v_i)| \le 1$, and the temperature hit rate the
fraction with $|\hat T_i - T_i| \le 25$\,°C. Hit rates are reported because
the true answer is one of a short list of settings. Coverage is the fraction
of true values inside the central 90\% predictive interval.

The baselines answer specific questions. The most frequent training alloy or
$(\Text, \vext)$ pair gives the score of a model that has learned nothing
about the input. The known-half-only baseline is a condition-balanced CatBoost
classifier that sees only the known process (Task A) or the known composition
(Task B), selected on the same validation conditions; it measures how much
the design confound alone reveals. Retrieval is the natural baseline when the
answer is a catalogue entry, GBDT and the ARD GP are standard regressors for
small tabular data and serve as the unconstrained counterparts of the
catalogue heads, and the plain classifiers of
Appendix~\ref{app:cls_ablation} keep the catalogue but drop the order of the
settings.

\section{Additional results}
\label{app:results}

Except for the classification ablation of Appendix~\ref{app:cls_ablation},
every number in this part is recomputed from the stored raw per-fold
predictions of the same condition-grouped 5-fold cross-validation used in the
main text, so no head is re-fitted. Figures that name a single fold use
\texttt{cv\_fold0} for illustration; the pooled figures aggregate all 107
held-out conditions.

\subsection{Per-fold tables}
\label{app:folds}

Table~\ref{tab:app_composition} shows that the ranking in
Table~\ref{tab:composition} is not an artefact of averaging. The conventional
reranker leads on all five folds, and its worst fold (20.4\% WAPE) is still
better than the best fold of either learned branch. The learned branches are
also far less stable: vision ranges from 28.0 to 73.8\% WAPE across folds,
which is why we quote fold means rather than any single split. The same fold
is the hardest for all three descriptors, so the difficulty is a property of
the held-out conditions rather than of one descriptor.

\begin{table}[t]
  \caption{Task A per fold: WAPE (\%) / alloy top-1 for each of the five
  held-out condition groups. Head names as in Table~\ref{tab:composition}.}
  \label{tab:app_composition}
  \centering\footnotesize
  \begin{tabularx}{\linewidth}{@{}llXXXXX@{}}
    \toprule
    repr. & head & fold 1 & fold 2 & fold 3 & fold 4 & fold 5 \\
    \midrule
    \multirow{3}{*}{conv.}
      & kNN retrieval & 38.1/0.35 & 20.0/0.63 & 26.6/0.50 & 28.2/0.57 & 37.2/0.43 \\
      & balanced fusion & 15.0/0.60 & 9.5/0.74 & 14.8/0.59 & 20.3/0.61 & 24.6/0.52 \\
      & reranker & 9.5/0.65 & 11.4/0.68 & 13.0/0.64 & 13.6/0.70 & 20.4/0.57 \\
    \midrule
    \multirow{3}{*}{vision}
      & kNN retrieval & 39.8/0.30 & 44.7/0.42 & 75.9/0.18 & 72.9/0.26 & 52.7/0.30 \\
      & balanced fusion & 31.9/0.40 & 47.0/0.32 & 28.0/0.32 & 42.9/0.22 & 73.8/0.13 \\
      & reranker & 31.9/0.40 & 31.2/0.37 & 28.0/0.32 & 42.9/0.22 & 73.8/0.13 \\
    \midrule
    \multirow{3}{*}{gnn}
      & kNN retrieval & 30.5/0.40 & 35.7/0.42 & 48.7/0.23 & 49.2/0.22 & 59.7/0.13 \\
      & balanced fusion & 31.9/0.35 & 43.9/0.32 & 46.4/0.27 & 45.4/0.22 & 54.1/0.17 \\
      & reranker & 31.9/0.35 & 43.9/0.32 & 45.4/0.27 & 45.4/0.22 & 54.1/0.17 \\
    \bottomrule
  \end{tabularx}
\end{table}

Table~\ref{tab:app_process} adds the standard deviations left out of
Table~\ref{tab:process} and scores velocity in log space alongside the raw
units. The joint-grid GP has the smallest spread across folds on nearly every
row, so its advantage is consistent rather than carried by one easy fold. The
ARD GP behaves in the opposite way: its velocity MAPE has a standard deviation
of the same order as its mean (120.5 $\pm$ 57.2 on vision), which is the
signature of a head whose error is dominated by a few conditions placed far
from any observed setting.

\begin{table}[t]
  \caption{Task B, full 5-fold statistics (mean $\pm$ std). $\log\vext$ rows
  score the same predictions in natural-log space. MAPE itself is undefined
  there because $\log\vext$ crosses zero inside the observed range, so those
  rows instead report $100 \times$ log-space MAE, the well-defined
  first-order approximation to percent error: 42 means the typical prediction
  is off by roughly 42\%.}
  \label{tab:app_process}
  \centering\small
  \begin{tabularx}{\linewidth}{@{}lllXX@{}}
    \toprule
    repr. & head & target & MAE & MAPE (\%) \\
    \midrule
    conv. & GBDT & $\Text$ & $41.6 \pm 10.5$ & $11.9 \pm 2.3$ \\
    conv. & ARD GP & $\Text$ & $52.1 \pm 6.9$ & $15.2 \pm 2.0$ \\
    conv. & joint-grid GP & $\Text$ & $38.1 \pm 5.6$ & $10.9 \pm 1.1$ \\
    conv. & GBDT & $\vext$ & $1.34 \pm 0.33$ & $64.5 \pm 15.6$ \\
    conv. & ARD GP & $\vext$ & $1.71 \pm 0.21$ & $104.7 \pm 40.9$ \\
    conv. & joint-grid GP & $\vext$ & $0.98 \pm 0.19$ & $48.7 \pm 12.4$ \\
    conv. & GBDT & $\log\vext$ & $0.580 \pm 0.069$ & $58.0 \pm 6.9$ \\
    conv. & ARD GP & $\log\vext$ & $0.751 \pm 0.138$ & $75.1 \pm 13.8$ \\
    conv. & joint-grid GP & $\log\vext$ & $0.420 \pm 0.083$ & $42.0 \pm 8.3$ \\
    \midrule
    vision & GBDT & $\Text$ & $39.2 \pm 10.1$ & $11.4 \pm 1.9$ \\
    vision & ARD GP & $\Text$ & $72.1 \pm 18.0$ & $19.9 \pm 3.4$ \\
    vision & joint-grid GP & $\Text$ & $35.2 \pm 7.3$ & $10.5 \pm 1.4$ \\
    vision & GBDT & $\vext$ & $1.22 \pm 0.32$ & $55.9 \pm 10.7$ \\
    vision & ARD GP & $\vext$ & $1.76 \pm 0.53$ & $120.5 \pm 57.2$ \\
    vision & joint-grid GP & $\vext$ & $1.10 \pm 0.24$ & $49.6 \pm 14.5$ \\
    vision & GBDT & $\log\vext$ & $0.491 \pm 0.051$ & $49.1 \pm 5.1$ \\
    vision & ARD GP & $\log\vext$ & $0.742 \pm 0.279$ & $74.2 \pm 27.9$ \\
    vision & joint-grid GP & $\log\vext$ & $0.450 \pm 0.109$ & $45.0 \pm 10.9$ \\
    \midrule
    gnn & GBDT & $\Text$ & $36.5 \pm 10.1$ & $10.7 \pm 2.2$ \\
    gnn & ARD GP & $\Text$ & $49.5 \pm 8.9$ & $14.7 \pm 3.1$ \\
    gnn & joint-grid GP & $\Text$ & $36.6 \pm 7.8$ & $10.8 \pm 1.9$ \\
    gnn & GBDT & $\vext$ & $1.11 \pm 0.33$ & $56.0 \pm 16.8$ \\
    gnn & ARD GP & $\vext$ & $1.44 \pm 0.16$ & $94.5 \pm 38.8$ \\
    gnn & joint-grid GP & $\vext$ & $1.01 \pm 0.26$ & $51.4 \pm 13.9$ \\
    gnn & GBDT & $\log\vext$ & $0.475 \pm 0.085$ & $47.5 \pm 8.5$ \\
    gnn & ARD GP & $\log\vext$ & $0.626 \pm 0.112$ & $62.6 \pm 11.2$ \\
    gnn & joint-grid GP & $\log\vext$ & $0.443 \pm 0.059$ & $44.3 \pm 5.9$ \\
    \bottomrule
  \end{tabularx}
\end{table}

\subsection{Calibration}
\label{app:calibration}

The joint-grid GP intervals are under-dispersed (90\% coverage $\approx$
0.5), because the grid decode concentrates mass on observed pairs. The plain
ARD GP on gnn is the best-calibrated head, with coverage 0.69 for temperature
and 0.71 for velocity. The winning heads are therefore point-accurate, and
their uncertainty statements should not be trusted beyond that.

\subsection{Classification ablation for Task B}
\label{app:cls_ablation}

If the process answer is discrete, a natural question is whether a plain
classifier should replace the regressor. We tested this with two classifiers
that receive exactly the Task B input of the other heads (descriptor, known
composition and extrusion-ratio flag) on the same folds. The \emph{pair
classifier} has one class per $(\Text, \vext)$ pair observed in the training
fold, 33 to 38 classes depending on the fold. The \emph{level classifiers} fit
one classifier over the temperature levels and a separate one over the speed
levels. Both are condition-balanced CatBoost classifiers on image-level rows,
with one of three configurations chosen on the validation conditions. Image
probabilities are averaged per condition and the most probable class is
returned. Table~\ref{tab:app_cls} compares them with the joint-grid GP and
with the two baselines that see no microstructure.

\begin{table}[t]
  \caption{Task B classification ablation, mean $\pm$ std over the five
  folds. T MAE in \textdegree C; exact and within-one are speed hit rates (\%).}
  \label{tab:app_cls}
  \centering\footnotesize
  \begin{tabularx}{\linewidth}{@{}llXXXXX@{}}
    \toprule
    repr. & head & T MAE & v MAPE (\%) & log MAE & exact v & within one \\
    \midrule
    \multirow{2}{*}{none}
      & most frequent pair & $84.3 \pm 4.2$ & $70.1 \pm 23.1$ & $1.050 \pm 0.204$ & $13.9 \pm 7.9$ & $20.2 \pm 8.2$ \\
      & composition only & $69.7 \pm 17.2$ & $94.9 \pm 35.7$ & $0.748 \pm 0.152$ & $22.5 \pm 7.5$ & $26.4 \pm 12.0$ \\
    \midrule
    \multirow{3}{*}{conv.}
      & pair classifier & $42.2 \pm 8.7$ & $92.8 \pm 46.0$ & $0.630 \pm 0.157$ & $27.0 \pm 6.3$ & $31.7 \pm 8.7$ \\
      & level classifiers & $37.9 \pm 8.4$ & $78.1 \pm 20.0$ & $0.610 \pm 0.040$ & $30.7 \pm 6.6$ & $34.6 \pm 6.2$ \\
      & joint-grid GP & $38.1 \pm 5.6$ & $48.7 \pm 12.4$ & $0.420 \pm 0.083$ & $24.0 \pm 5.6$ & $51.8 \pm 12.9$ \\
    \midrule
    \multirow{3}{*}{vision}
      & pair classifier & $40.9 \pm 11.5$ & $63.0 \pm 10.2$ & $0.567 \pm 0.084$ & $27.0 \pm 7.4$ & $35.6 \pm 6.3$ \\
      & level classifiers & $43.4 \pm 9.9$ & $78.9 \pm 13.8$ & $0.618 \pm 0.046$ & $26.8 \pm 4.0$ & $29.8 \pm 3.8$ \\
      & joint-grid GP & $35.2 \pm 7.3$ & $49.6 \pm 14.5$ & $0.450 \pm 0.109$ & $24.3 \pm 14.8$ & $48.3 \pm 13.1$ \\
    \midrule
    \multirow{3}{*}{gnn}
      & pair classifier & $35.0 \pm 7.5$ & $67.5 \pm 22.2$ & $0.601 \pm 0.123$ & $23.2 \pm 7.8$ & $29.1 \pm 8.2$ \\
      & level classifiers & $42.0 \pm 9.9$ & $73.8 \pm 15.7$ & $0.587 \pm 0.071$ & $25.0 \pm 6.2$ & $28.9 \pm 4.1$ \\
      & joint-grid GP & $36.6 \pm 7.8$ & $51.4 \pm 13.9$ & $0.443 \pm 0.059$ & $28.8 \pm 9.4$ & $46.0 \pm 6.9$ \\
    \bottomrule
  \end{tabularx}
\end{table}

The classifiers name the exact speed about as often as the joint-grid GP, and
the conventional level classifiers slightly more often (30.7 against 24.0\%).
Their misses, however, are large. A classifier has no notion that neighbouring
settings are close, so a wrong answer is as likely to be far from the truth
as next to it. Their share within one setting (29 to 36\%) is barely above the
composition-only baseline (26\%), and their log speed error (0.57 to 0.63)
lies between that baseline and the joint-grid GP. On temperature the
classifiers are competitive, and on the gnn branch the pair classifier has the
lowest temperature MAE of any head (35.0\,\textdegree C). The ablation
therefore sharpens the main-text conclusion rather than overturning it:
treating the answer as discrete helps when the head chooses among the observed
settings and also keeps the distances between them, which is what the
joint-grid GP does.

\subsection{Composition diagnostics}
\label{app:comp_diag}

\begin{figure}[t]
  \centering
  \includegraphics[width=0.99\linewidth]{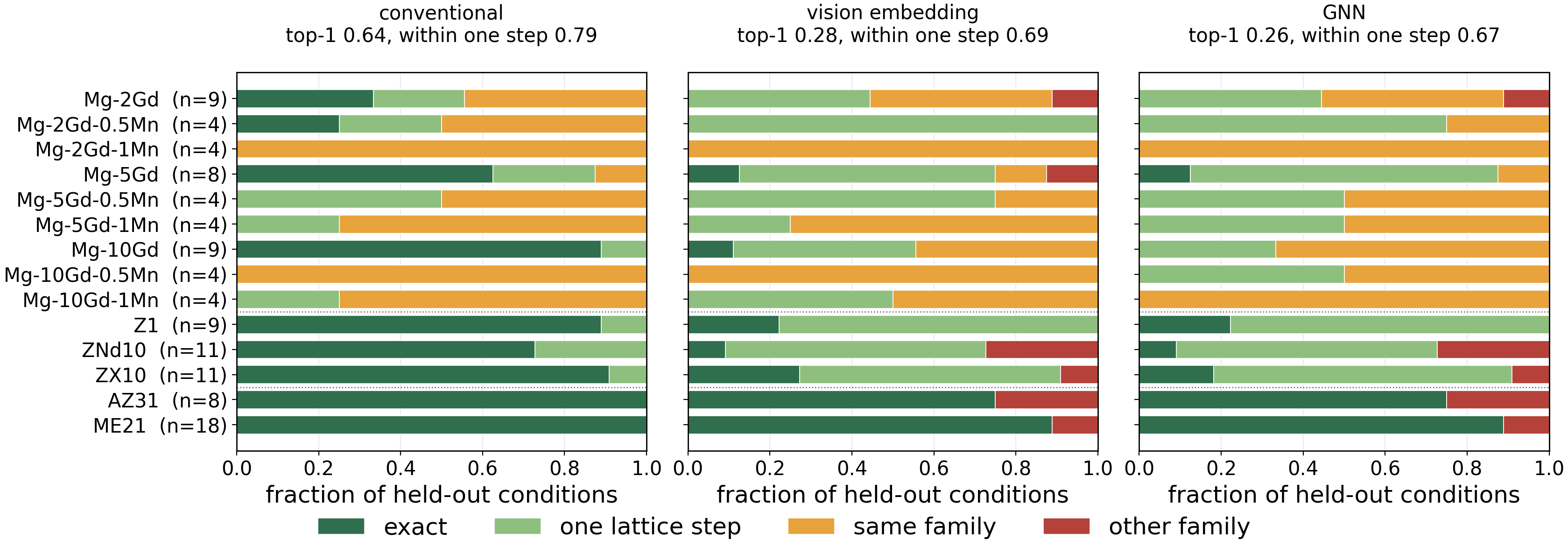}
  \caption{Where the top-1 alloy call lands, per alloy, pooled over all 107
  held-out conditions. One step is one Gd or one Mn increment inside the
  Mg-Gd(-Mn) grid, or a swap within the Z1 / ZNd10 / ZX10 trio.}
  \label{fig:app_taxonomy}
\end{figure}

\begin{figure}[t]
  \centering
  \includegraphics[width=0.99\linewidth]{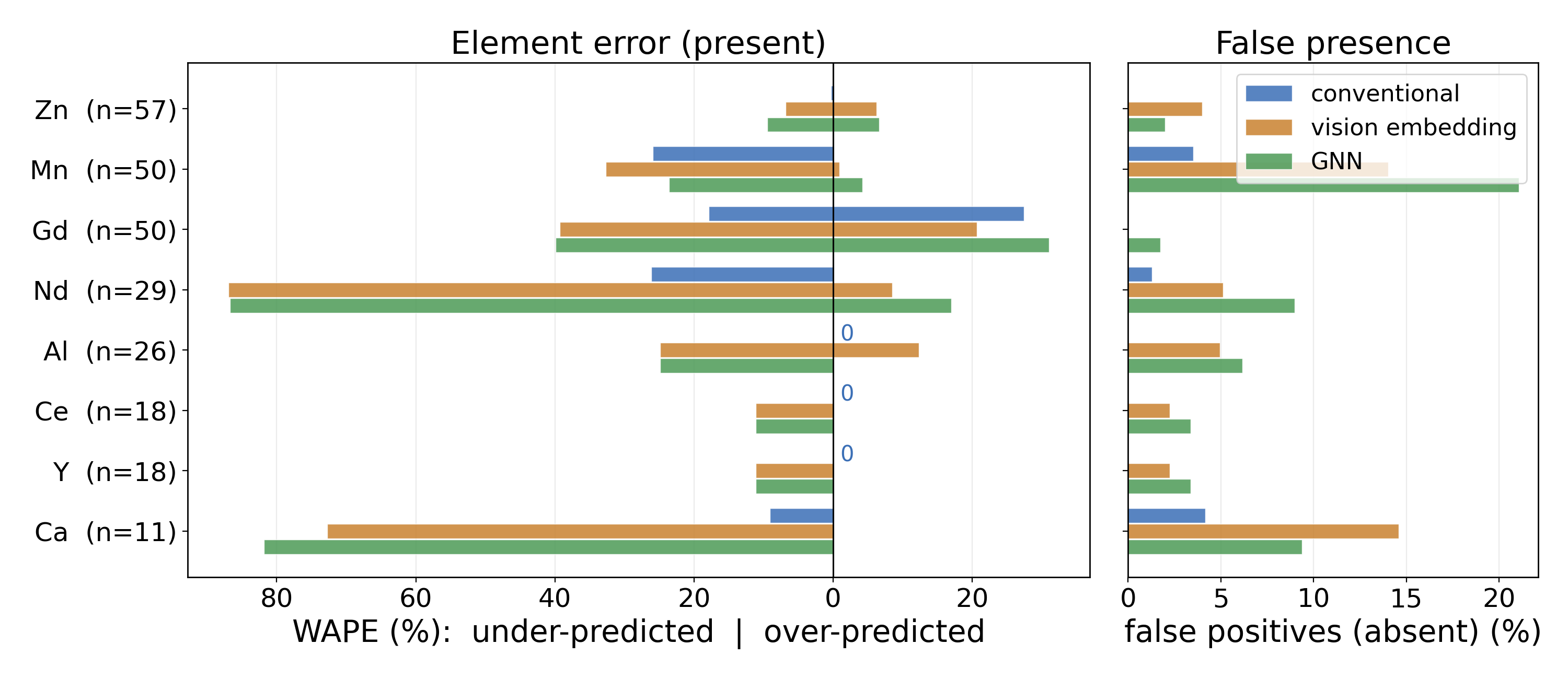}
  \caption{Per-element error. Left: WAPE over the conditions that contain the
  element, split at zero into the under-predicted and over-predicted share,
  where ``0'' marks a branch exactly right on every such condition. Right: the
  presence false-positive rate.}
  \label{fig:app_elements}
\end{figure}

The WAPE in Table~\ref{tab:composition} mixes two different questions: whether
the head recognises the right alloy system, and whether it resolves the
concentration inside that system. Figures~\ref{fig:app_taxonomy} and
\ref{fig:app_elements} separate them.

Figure~\ref{fig:app_taxonomy} sorts each top-1 call by how far it lands from
the true alloy. The conventional branch is exact on AZ31 and ME21, and reaches
0.79 within one composition step against a top-1 of 0.64, so most of its
mistakes are a neighbouring Gd or Mn level rather than an unrelated alloy. The
two learned branches sit near 0.68 within one step while their top-1 stays
below 0.30, meaning their mistakes are spread across the catalogue instead of
being concentrated on close neighbours. The alloy rows hold between 4 and 18
conditions, so a single misclassified condition moves a small row by about
25\%; the pattern across rows is the robust part.

Figure~\ref{fig:app_elements} localises the same effect per element. The
conventional branch is exact on Zn, Al, Ce and Y, and its entire error sits in
Gd, Mn, Nd and Ca, which are the elements that vary in small increments inside
the Mg-Gd(-Mn) family. The learned branches under-predict the rare elements Nd
and Ca and simultaneously add Mn and Ca to alloys that do not contain them,
which is what the false-positive column of Table~\ref{tab:composition} counts.
Because the head always outputs one of the 14 known alloys, the per-element
error can take only a few distinct values, so the figure uses bars rather than
a density. Taken together, the two figures say that the gap between the
descriptors is a gap in concentration resolution, not in chemistry
recognition.

\subsection{Process diagnostics}
\label{app:proc_diag}

\begin{figure}[t]
  \centering
  \includegraphics[width=0.99\linewidth]{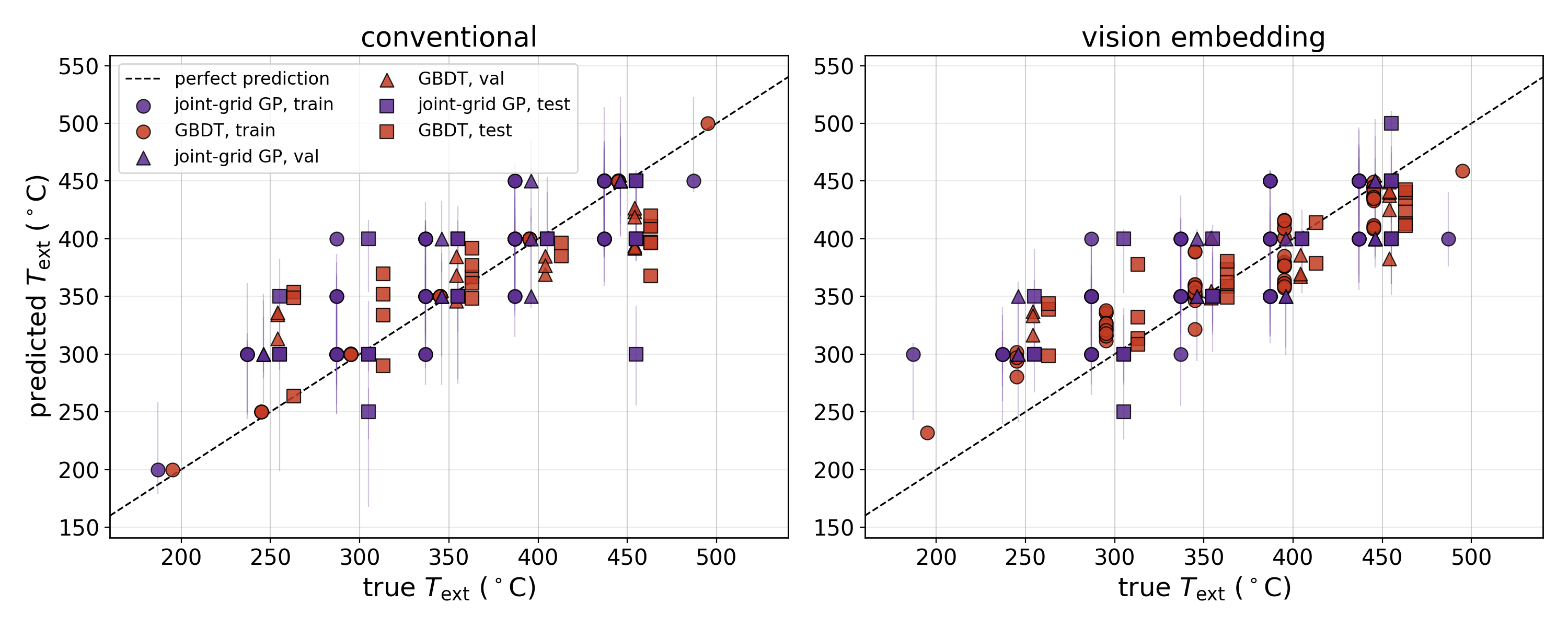}
  \caption{Task B temperature: predicted against true $\Text$
  (\texttt{cv\_fold0}). Purple = joint-grid GP (bar = latent 90\% interval),
  red = GBDT; circles / triangles / squares = train / val / test, dashed =
  perfect prediction.}
  \label{fig:temp_parity}
\end{figure}

Section~\ref{sec:results} argued that the process error is tail shrinkage
carried by the descriptor rather than by the head. For temperature,
Figure~\ref{fig:temp_parity} shows the predicted-versus-true view: both heads
track the diagonal but saturate above $\approx 400$\degC, and the joint-grid
GP's latent intervals cover the diagonal on most held-out conditions.

\begin{figure}[t]
  \centering
  \includegraphics[width=0.99\linewidth]{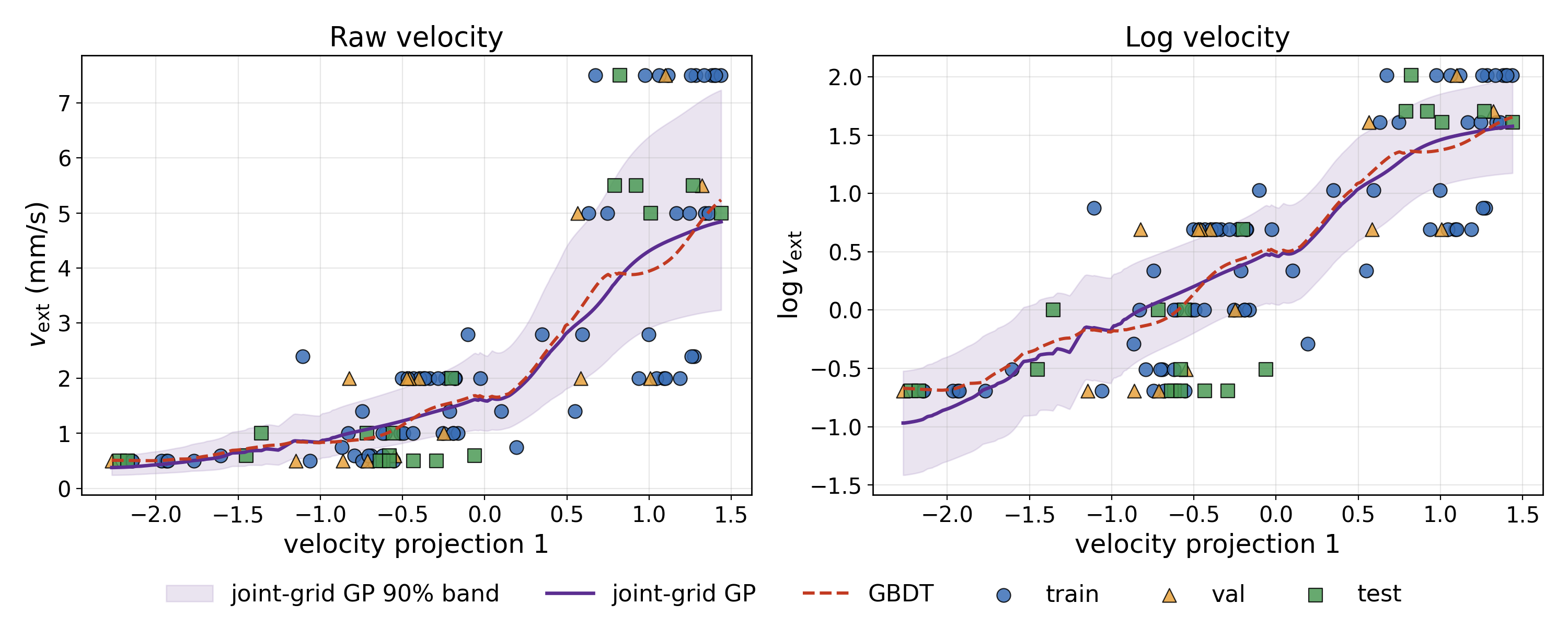}
  \caption{Velocity response along a supervised 1-D projection on the GNN
  branch, raw (left) and log (right).}
  \label{fig:gnn_velocity}
\end{figure}

For velocity, Figure~\ref{fig:gnn_velocity} gives the response along a
supervised 1-D projection on the GNN branch. It is a sigmoid that saturates at
the fast end, the velocity counterpart of Figure~\ref{fig:lda_temperature}.

\begin{figure}[t]
  \centering
  \includegraphics[width=0.99\linewidth]{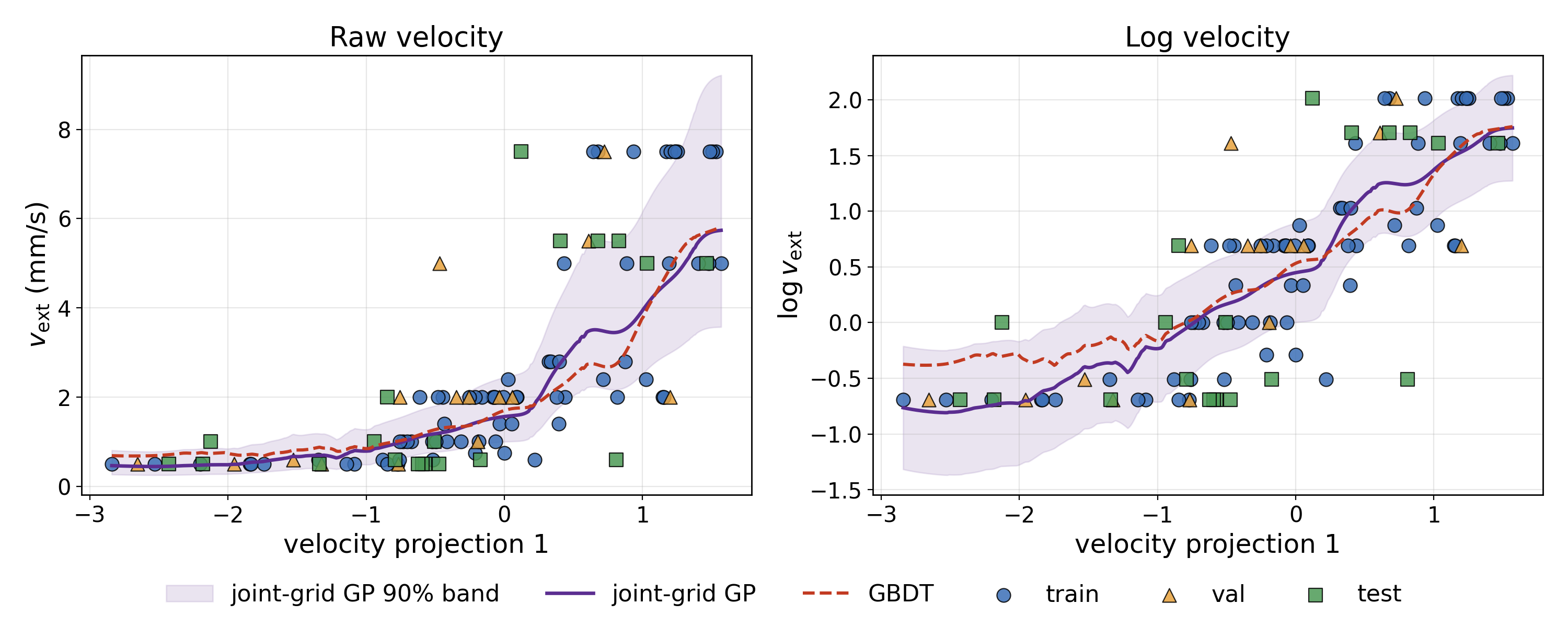}
  \caption{Velocity response along a supervised 1-D projection on the
  conventional branch, raw (left) and log (right).}
  \label{fig:app_conv_velocity}
\end{figure}

\begin{figure}[t]
  \centering
  \includegraphics[width=0.99\linewidth]{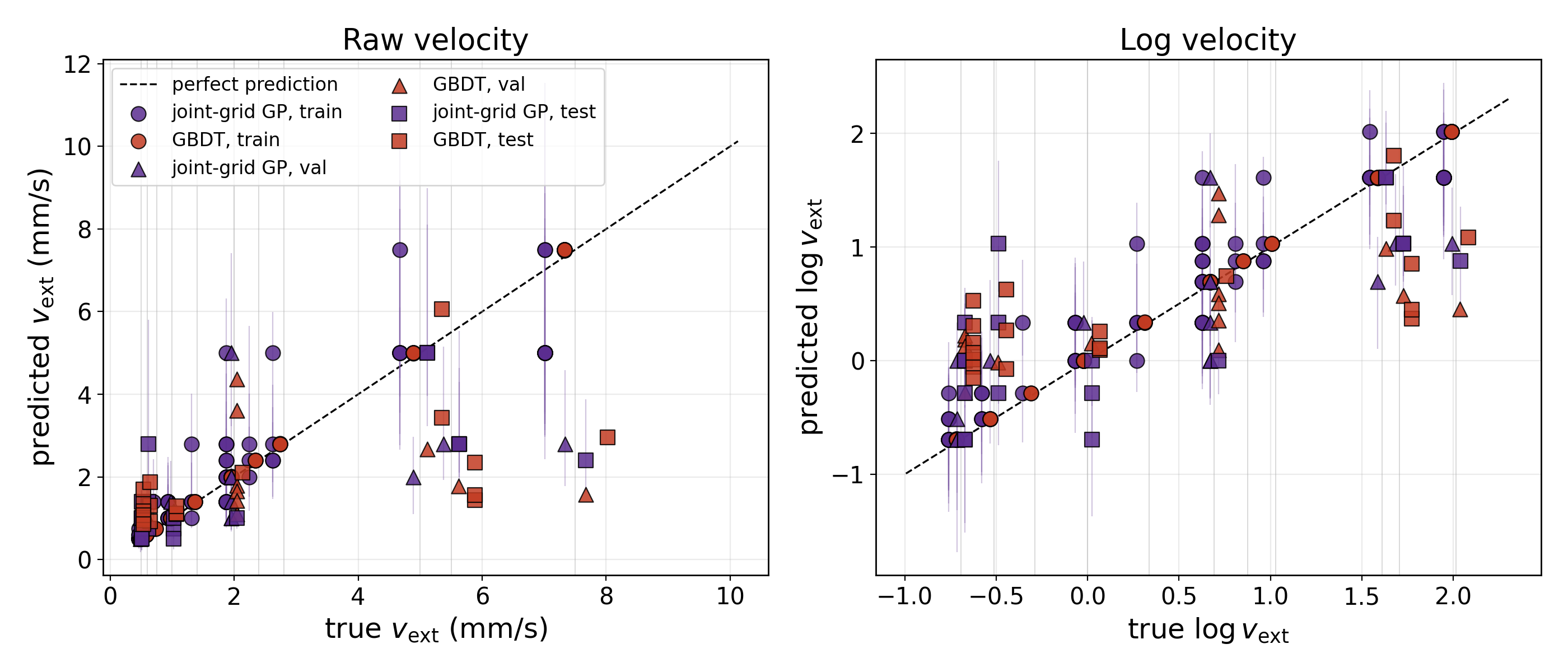}
  \caption{Velocity parity on the conventional branch (\texttt{cv\_fold0}),
  raw (left) and log (right).}
  \label{fig:app_conv_vel_parity}
\end{figure}

\begin{figure}[t]
  \centering
  \includegraphics[width=0.99\linewidth]{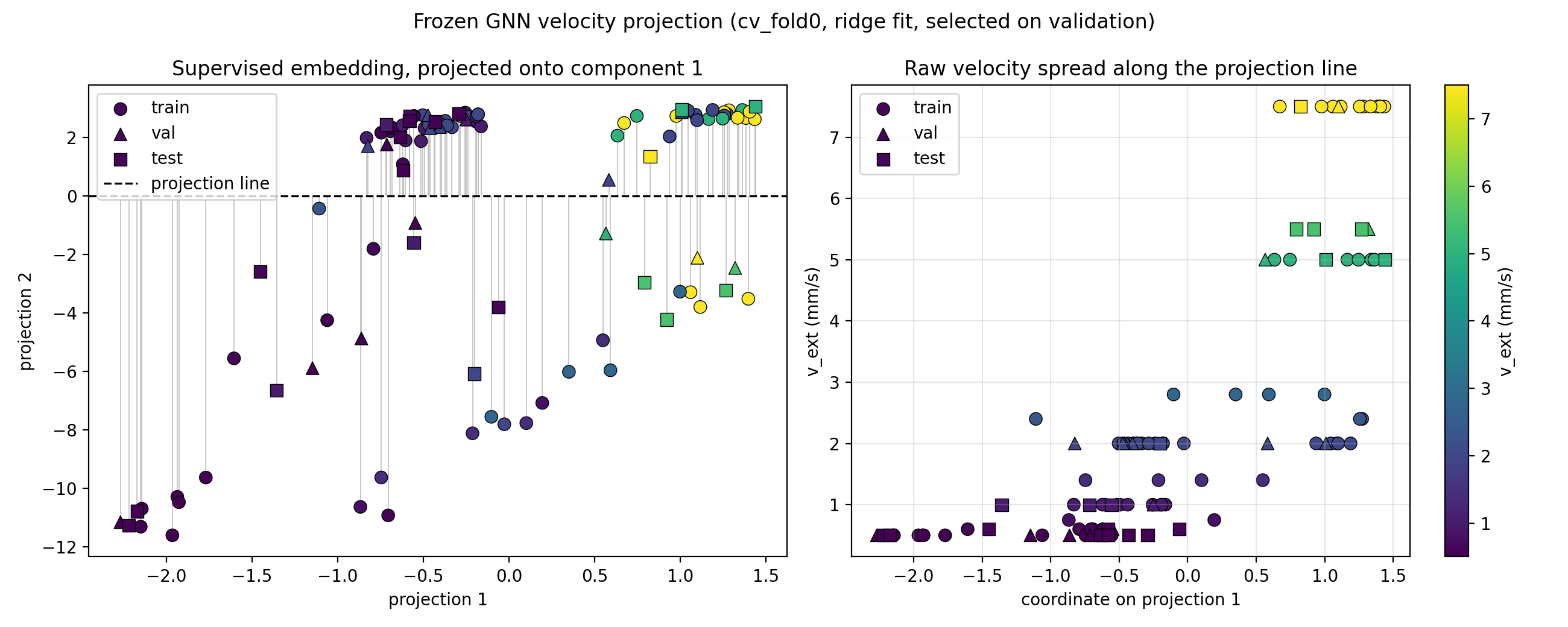}
  \caption{The frozen GNN velocity projection (\texttt{cv\_fold0}). Left: the
  2-D supervised embedding, points coloured by true $\vext$ and joined to the
  projection line. Right: the raw velocity spread along component 1.}
  \label{fig:app_gnn_projection}
\end{figure}

Figures~\ref{fig:app_conv_velocity} and \ref{fig:app_conv_vel_parity} repeat
the GNN analysis of Figures~\ref{fig:gnn_velocity} and \ref{fig:vel_parity}
on the conventional branch, and Figure~\ref{fig:app_gnn_projection} shows the
2-D embedding that the 1-D response curves are read from.

The conclusion holds on both branches. The response along the supervised
projection rises monotonically with speed and then flattens at the fast end,
so conditions that differ only in speed among the fastest settings are barely
distinguishable in the descriptor, whichever head reads it. The conventional
projection is the less well behaved of the two, because a single extreme
low-velocity cluster near $-3$ dominates it, yet its saturation is the same.

The parity view on the conventional branch repeats what the GNN branch showed
in Figure~\ref{fig:vel_parity}: predictions are tight at low speed and
undershoot at high speed.

The 2-D projection makes the mechanism explicit. The slow conditions spread
out along the embedding while the fast ones are compressed into a narrow band,
and that geometry is what produces a sigmoidal 1-D response and predictions
pulled toward the middle of the speed range.

\end{document}